\pdfoutput=1

\documentclass[11pt]{article}

\usepackage[final]{acl}

\usepackage{times}
\usepackage{latexsym}

\usepackage[T1]{fontenc}

\usepackage[utf8]{inputenc}
\usepackage{CJKutf8}

\usepackage{microtype}

\usepackage{inconsolata}

\usepackage{graphicx}

\usepackage{amsmath}

\definecolor{codegray}{rgb}{0.5,0.5,0.5}
\definecolor{codepurple}{rgb}{0.58,0,0.82}
\definecolor{backcolour}{rgb}{0.95,0.95,0.92}
\definecolor{deepgreen}{rgb}{0.0, 0.5, 0.0}

\usepackage{booktabs} 
\usepackage{makecell} 

\usepackage{caption} 

\usepackage{colortbl}
\definecolor{customblue}{HTML}{74AED4} 
\definecolor{customgreen}{HTML}{D3E2B7} 
\definecolor{customred}{HTML}{ECA8A9} 
\definecolor{custompurple}{HTML}{CFAFD4}
\definecolor{customorange}{HTML}{FFD599}

\definecolor{tableorange}{HTML}{FFD599}
\definecolor{tablepurple}{HTML}{E3D4E7}
\definecolor{tableblue}{HTML}{D9E8FA}
\definecolor{tablegreen}{HTML}{D6E9D6}

\usepackage[most]{tcolorbox}
\usepackage{float}
\usepackage{xspace}
\tcbset{
  aibox/.style={
    width=\textwidth,
    top=10pt,
    colback=white,
    colframe=black,
    colbacktitle=black,
    enhanced,
    center,
    attach boxed title to top left={yshift=-0.1in,xshift=0.15in},
    boxed title style={boxrule=0pt,colframe=white,},
  }
}
\newtcolorbox{AIbox}[2][]{aibox,title=#2,#1}
\definecolor{darkgreen}{rgb}{0.0, 0.5, 0.0}
\definecolor{darkgray}{gray}{0.4}
\definecolor{maroon}{rgb}{0.5, 0.0, 0.0}
\definecolor{navy}{rgb}{0.0, 0.0, 0.5}
\definecolor{teal}{rgb}{0.0, 0.5, 0.5}

\usepackage{pifont} 
\usepackage{textcomp} 

\title{Exploring Compositional Generalization of Multimodal LLMs\\for Medical Imaging}

\author{
Zhenyang Cai$^\dagger$,
Junying Chen$^\dagger$,
Rongsheng Wang$^\dagger$,
Weihong Wang, \\
\textbf{Yonglin Deng, Dingjie Song, Yize Chen, Zixu Zhang,}
\textbf{Benyou Wang}$^*$ \\
The Chinese University of Hong Kong, Shenzhen \\
\textit{wangbenyou@cuhk.edu.cn}\\
}

\begin{document}
\begin{CJK}{UTF8}{gkai}
\maketitle


\renewcommand{\thefootnote}{\fnsymbol{footnote}}
\footnotetext[2]{Equal Contribution. $^*$Corresponding author.}
\renewcommand{\thefootnote}{\arabic{footnote}}


\begin{abstract}

Medical imaging provides essential visual insights for diagnosis, and multimodal large language models (MLLMs) are increasingly utilized for its analysis due to their strong generalization capabilities; however, the underlying factors driving this generalization remain unclear. Current research suggests that multi-task training outperforms single-task as different tasks can benefit each other, but they often overlook the internal relationships within these tasks. To analyze this phenomenon, we attempted to employ \textit{compositional generalization} (CG), which refers to the models' ability to understand novel combinations by recombining learned elements, as a guiding framework. Since medical images can be precisely defined by \textbf{M}odality, \textbf{A}natomical area, and \textbf{T}ask, naturally providing an environment for exploring CG, we assembled 106 medical datasets to create \textbf{Med-MAT} for comprehensive experiments. The experiments confirmed that MLLMs can use CG to understand unseen medical images and identified CG as one of the main drivers of the generalization observed in multi-task training. Additionally, further studies demonstrated that CG effectively supports datasets with limited data and confirmed that MLLMs can achieve CG across classification and detection tasks, underscoring its broader generalization potential. Med-MAT is available at \href{https://github.com/FreedomIntelligence/Med-MAT}{https://\\github.com/FreedomIntelligence/Med-MAT}.
\end{abstract}


\begin{figure}[ht!]
    \centering
    \includegraphics[width=0.48\textwidth]{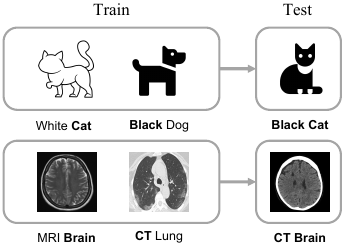}
    \caption{Examples of \textit{Compositional Generalization}: The model is required to understand unseen images by recombining the fundamental elements it has learned.}
    \label{fig:CG}
    \vspace{-6mm}
\end{figure}

\begin{figure*}[ht!]
    \centering
    \includegraphics[width=\textwidth]{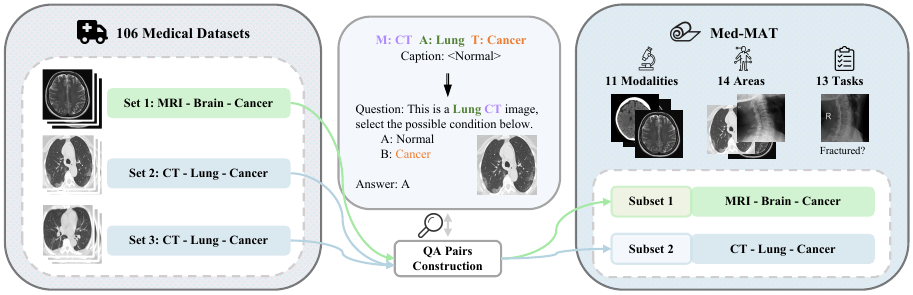}
    \caption{The process of integrating a vast amount of labeled medical image data to create Med-MAT.}
    \label{fig:Med-MAT}
    \vspace{-4mm}
\end{figure*}

\section{Introduction}

Medical imaging provides essential visual insights into the structures of the human body, making it a critical tool for medical diagnosis. Recently, multimodal large language models (MLLMs)~\cite{liu2023llava, li2024llava, chen2024huatuogpt} have been employed to analyze these images due to their strong interpretability and generalization capabilities. 
In this paper, we focus on the latter: generalization of MLLMs in medical imaging.  
Current research \cite{mo2024multimed, ren2024medical} has demonstrated that models trained on multiple tasks outperform those trained on a single task as they can leverage potential knowledge from other tasks.
Yet, the underlying factors that contribute to this generalization remain insufficiently explored.

To this end,  we take the perspective of \textit{composition generalization }(CG)~\cite{li2019compositional, xu2022compositional, tang2024sparkle} to investigate the generalization phenomenon of mutual improvement in MLLMs' understanding of medical images. Specifically, CG is the model's ability to learn fundamental elements and recombine them in novel ways to understand unseen combinations (e.g., learning \textit{\textbf{Cat}} from \textit{White \textbf{Cat}} and \textit{\textbf{Black}} from \textit{\textbf{Black} Dog}, then generalizing to \textit{\textbf{Black} \textbf{Cat}}, as shown in Figure \ref{fig:CG}).

In this paper,  we categorize each image to three elements: \textbf{M}odality\includegraphics[height=0.8em]{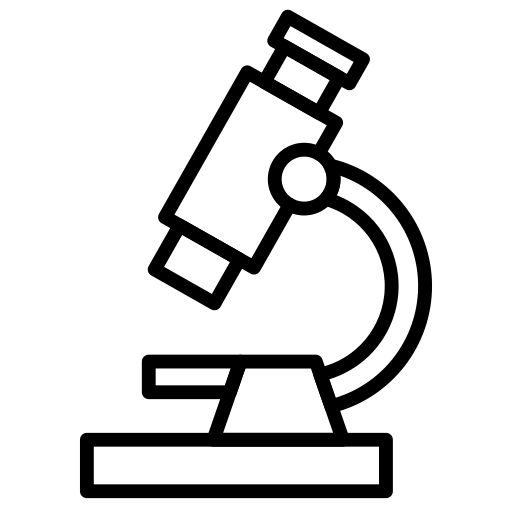}, \textbf{A}natomical area\includegraphics[height=0.8em]{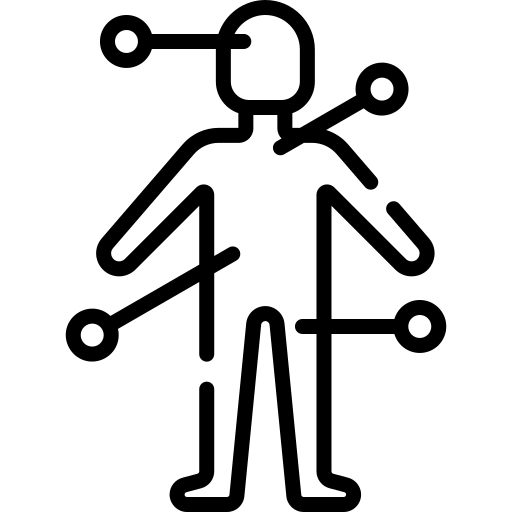}, and medical \textbf{T}ask\includegraphics[height=0.8em]{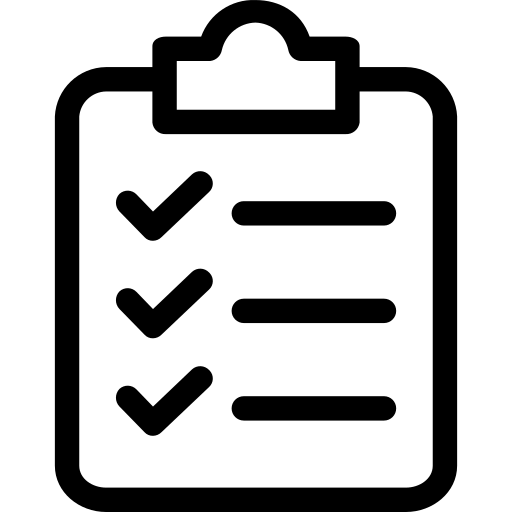}, presenting numerous natural opportunities for CG. We defined these three elements as the \textbf{MAT-Triplet} and collected 106 medical datasets, subsequently merging those that share the same \textit{MAT-Triplet} to create the \textbf{Med-MAT} dataset. Ultimately, Med-MAT comprises 53 subsets, encompassing 11 modalities, 14 anatomical regions, and 13 medical tasks, providing a foundation for investigating CG and other generalization methods.

To verify the existence of CG, we designated specific datasets as \textit{\textbf{Target}} data and selected all \textit{\textbf{Related}} data from Med-MAT that shared the same MAT-Triplet with the \textit{Target} data. Using these data combinations, we accessed the generalization performance of MLLMs and observed that they could leverage CG to understand unseen medical images. To further validate this finding, we repeated the experiments on different MLLMs and obtained consistent results, confirming the universality of CG.

Building on these insights, we expanded the number of combinations and observed the changes in model generalization performance after deliberately disrupting CG, ultimately revealing that CG is a key factor driving the generalization of MLLMs. Furthermore, we explored the potential applications of CG and its performance across classification and detection tasks, finding that CG enhances MLLMs' ability to handle medical scenarios with limited training data and improves their capacity for spatial awareness.

Here are the key contributions of our work: 1) A VQA dataset, Med-MAT, has been constructed, providing a platform to explore the generalization of MLLMs on medical images. 2) Through this dataset, we observed that MLLMs in different architectures can utilize compositional generalization to understand unseen images and demonstrated that this is one of the main forms of generalization for medical MLLMs. 3) Finally, the real-world applicability of CG, along with its presence across detection and classification tasks, has been further explored, highlighting its potential to enhance data-efficient training and its broad applicability.

\begin{table*}[ht!]
\setlength{\tabcolsep}{2pt}
\centering
\small
\begin{tabular}{l|rrrrrrrrrrrrrrrrrrrrrrrrr}
\toprule
\textbf{Subset No.} & 02 & 03 & 07 & 08 & 09 & 11 & 13 & 14 & 15 & 16 & 18 & 19 & 21 & 22 & 23 & 25 & 26 & 28 & 30 & 31 & 32 & 33 & 35 & 36 & 37 \\
\midrule
\textit{Baseline} & 21 & 47 & 40 & 25 & 26 & 27 & 28 & 24 & 22 & 24 & 25 & 23 & 49 & 26 & 25 & 24 & 49 & 30 & 49 & 21 & 49 & 20 & 25 & 23 & 19 \\
\textit{Single-task Training} & \underline{24} & \underline{49} & \underline{50} & \underline{68} & \underline{65} & \underline{76} & \underline{83} & \underline{53} & \underline{61} & \underline{32} & \underline{29} & \underline{26} & \underline{57} & \underline{53} & \underline{28} & \underline{24} & \underline{57} & \underline{64} & \underline{89} & \underline{60} & \underline{97} & \underline{54} & \underline{29} & \underline{51} & \underline{49} \\
\textit{Multi-task Training} & \textbf{96} & \textbf{89} & \textbf{80} & \textbf{80} & \textbf{79} & \textbf{97} & \textbf{92} & \textbf{88} & \textbf{76} & \textbf{57} & \textbf{88} & \textbf{74} & \textbf{87} & \textbf{86} & \textbf{93} & \textbf{52} & \textbf{98} & \textbf{72} & \textbf{94} & \textbf{61} & \textbf{100} & \textbf{72} & \textbf{75} & \textbf{60} & \textbf{50} \\
\bottomrule
\end{tabular}
\caption{Accuracy(\%) of different models on In-Distribution datasets (each dataset contains over 3,000 samples, with 3,000 selected for training). Within each segment, \textbf{bold} highlights the best scores, and \underline{underline} indicates the second-best. \textit{Baseline} represents the results without any training, \textit{Single-task Training} refers to the results after training on a single dataset, and \textit{Multi-task Training} represents the results after training on all datasets.}
\label{tab:id-res}
\end{table*}

\begin{table*}[ht!]
\setlength{\tabcolsep}{2pt}
\centering
\small
\begin{tabular}{l|rrrrrrrrrrrr}
\toprule
\textbf{Subset No.} & 01 & 04 & 05 & 06 & 10 & 12 & 17 & 20 & 24 & 27 & 29 & 34\\
\midrule
\textit{Baseline} & 32 & 25 & 33 & \textbf{33} & 48 & 27 & 33 & 13 & 34 & 37 & 31 & 20 \\
\textit{Multi-task Training} & \textbf{39} & \textbf{26} & \textbf{70} & 31 & \textbf{58} & \textbf{38} & \textbf{61} & \textbf{40} & \textbf{35} & \textbf{41} & \textbf{55} & \textbf{50} \\
\bottomrule
\end{tabular}
\caption{Accuracy(\%) of different models on Out-Of-Distribution Dataset (each dataset contains fewer than 3,000 samples and is used only for testing). \textbf{Bold} highlights the best scores. \textit{Multi-task Training} represents the results after training on all datasets.}
\label{tab:ood-res}
\vspace{-3mm}
\end{table*}

\section{A Pilot Study on Generalization}


\subsection{Data Collection (Med-MAT)}
\label{sec:matsec1}

Most existing datasets for MLLMs~\cite{zhang2023pmc, li2024llava, chen2024huatuogpt}, primarily VQA datasets, provide broad coverage but lack attribute annotations for individual samples, which are not suitable for CG exploration. To address this gap, we curated a large collection of image-text pairs to develop \textbf{Med-MAT}, ensuring that each sample is explicitly defined by MAT-Triplet.

\paragraph{Data Construction} Med-MAT contains a total of 106 image-label pair medical datasets, sourced from various medical public challenges or high-quality annotated datasets. All datasets are categorized according to their MAT-Triplet, with data having identical elements grouped into a single subset (Figure \ref{fig:Med-MAT}). Labels are manually clustered to ensure that annotations with the same meaning are not repeatedly used. In total, Med-MAT covers 11 medical modalities~\includegraphics[height=0.8em]{images/icon-modality.png}, 14 anatomical areas~\includegraphics[height=0.8em]{images/icon-area.png}, and 13 medical tasks~\includegraphics[height=0.8em]{images/icon-task.png}, hoping that it can spread across various medical tasks like a mat. (Data lists are shown in Appendix~\ref{app:dataset})

\paragraph{Data Distribution} All subsets are divided into training and test sets following their original distributions or using a 9:1 ratio. To ensure a fair comparison, each training set is limited to 3,000 samples~\footnote{Most datasets contain around 3,000 samples.}, with label balance maintained as much as possible. Any subset that cannot meet this requirement is treated as an OOD (out-of-distribution) dataset. For the test sets, we strictly balance the number of samples per label to ensure that the accuracy metric reliably reflects model performance.

\begin{figure}[ht!]
    \vspace{-2mm}
    \centering
    \includegraphics[width=0.45\textwidth]{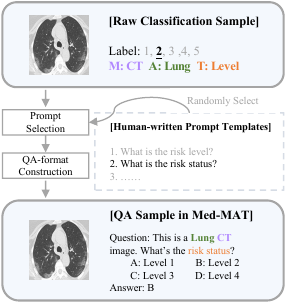}
    \caption{An example of formatting a raw classification sample into a Question-answering sample in Med-MAT.}
    \label{fig:qa-format}
    \vspace{-5mm}
\end{figure}

\paragraph{QA Pairs Construction} To enable MLLMs to directly train and test on Med-MAT, all image-label paired data were converted into a visual question-answering (VQA) format (Figure \ref{fig:qa-format}). Specifically, each subset was manually assigned 6 instructions to guide the MLLM in answering the subset task. For convenience, all samples were converted into single-choice questions with up to four options, and the remaining distractor options were randomly drawn from other labels within the subset. To mitigate potential evaluation biases arising from varying option counts, the ImageWikiQA dataset~\cite{zhang2024visually}, a non-medical dataset consisting of single-answer, four-option questions, was incorporated during the training.

\subsection{Observation}
\label{sec:matsec2}
\textbf{Experiment Setup} We chose LLaVA-v1.5-7B-Vicuna~\cite{liu2023llava} as the base model due to its transparent pretraining process and minimal use of medical data, reducing the risk of knowledge leakage. Leveraging MLLM's flexibility, we enabled task switching and generalization by adjusting prompts, streamlining generalization studies. Each experiment ran for 5 epochs on 8 A800 GPUs with a batch size of 32 and a learning rate of 5e-6.


\paragraph{Analysis} To access the generalization of MLLMs, we trained the baseline on all ID datasets to simulate \textit{Multi-task Training} and separately trained on individual ID datasets to establish the \textit{Single-task Training} as the control group. We then evaluated the models on all datasets. The results in Table \ref{tab:id-res} and \ref{tab:ood-res} confirm that \textit{Multi-task Training} outperformed \textit{Single-task Training} on specific tasks and improved OOD prediction, suggesting certain data combinations enhance classification and identifying valuable combinations for medical tasks warrants further research. This observation leads to a research question (\textbf{\textit{RQ}}):

\begin{quote}
\vspace{-2mm}
\textit{What drives the generalization observed in MLLMs during Multi-task Training?}
\vspace{-2mm}
\end{quote}

To address it, we aim to explore the generalization mechanism of MLLMs from the perspective of compositional generalization (CG).




\section{Proof of Concept on CG}

This section will prove the existence of CG in MLLMs, offering preliminary insights to address the \textbf{\textit{RQ}} and providing support for further analysis.


\begin{table*}[ht!]
\setlength{\tabcolsep}{2pt}
\centering
\small
\resizebox{0.88\textwidth}{!}{
\begin{tabular}{ll|ll|ll|ccc|c}
\toprule
\multicolumn{4}{c}{\textbf{Related Combination}} &  \multicolumn{2}{c}{\textbf{Target Subset}} & \textbf{Baseline} & \textbf{Baseline+} & \textbf{Trained} & \textbf{CG Helps} \\
\midrule
\includegraphics[height=0.8em]{images/icon-area.png}\textbf{Lung} & \includegraphics[height=0.8em]{images/icon-task.png}COVID & \includegraphics[height=0.8em]{images/icon-area.png}Brain & \includegraphics[height=0.8em]{images/icon-task.png}\textbf{Cancer} & \includegraphics[height=0.8em]{images/icon-area.png}\textbf{Lung} & \includegraphics[height=0.8em]{images/icon-task.png}\textbf{Cancer} & 25 & 25 & 27 & \ding{51} \\
\includegraphics[height=0.8em]{images/icon-area.png}\textbf{Lung} & \includegraphics[height=0.8em]{images/icon-task.png}Cancer & \includegraphics[height=0.8em]{images/icon-area.png}Brain & \includegraphics[height=0.8em]{images/icon-task.png}\textbf{State} & \includegraphics[height=0.8em]{images/icon-area.png}\textbf{Lung} & \includegraphics[height=0.8em]{images/icon-task.png}\textbf{State} & 47 & 46 & 50  & \ding{51} \\
\includegraphics[height=0.8em]{images/icon-area.png}\textbf{Brain} & \includegraphics[height=0.8em]{images/icon-task.png}Cancer & \includegraphics[height=0.8em]{images/icon-area.png}Lung & \includegraphics[height=0.8em]{images/icon-task.png}\textbf{State} & \includegraphics[height=0.8em]{images/icon-area.png}\textbf{Brain} & \includegraphics[height=0.8em]{images/icon-task.png}\textbf{State} & 33 & 50 & 57 & \ding{51} \\
\includegraphics[height=0.8em]{images/icon-area.png}\textbf{Bones} & \includegraphics[height=0.8em]{images/icon-task.png}Level & \includegraphics[height=0.8em]{images/icon-area.png}Lung & \includegraphics[height=0.8em]{images/icon-task.png}\textbf{State} & \includegraphics[height=0.8em]{images/icon-area.png}\textbf{Bones} & \includegraphics[height=0.8em]{images/icon-task.png}\textbf{State} & 49 & 53 & 51 & \ding{55} \\
\includegraphics[height=0.8em]{images/icon-area.png}\textbf{Bones} & \includegraphics[height=0.8em]{images/icon-task.png}Level & \includegraphics[height=0.8em]{images/icon-area.png}Brain & \includegraphics[height=0.8em]{images/icon-task.png}\textbf{State} & \includegraphics[height=0.8em]{images/icon-area.png}\textbf{Bones} & \includegraphics[height=0.8em]{images/icon-task.png}\textbf{State} & 49 & 53 & 72 & \ding{51} \\
\includegraphics[height=0.8em]{images/icon-area.png}\textbf{Bones} & \includegraphics[height=0.8em]{images/icon-task.png}Level & \includegraphics[height=0.8em]{images/icon-area.png}Breast & \includegraphics[height=0.8em]{images/icon-task.png}\textbf{Diseases} & \includegraphics[height=0.8em]{images/icon-area.png}\textbf{Bones} & \includegraphics[height=0.8em]{images/icon-task.png}\textbf{Diseases} & 37 & 33 & 39 & \ding{51} \\
\includegraphics[height=0.8em]{images/icon-area.png}\textbf{Bones} & \includegraphics[height=0.8em]{images/icon-task.png}Level & \includegraphics[height=0.8em]{images/icon-area.png}Lung & \includegraphics[height=0.8em]{images/icon-task.png}\textbf{Diseases} & \includegraphics[height=0.8em]{images/icon-area.png}\textbf{Bones} & \includegraphics[height=0.8em]{images/icon-task.png}\textbf{Diseases} & 37 & 33 & 43 & \ding{51} \\
\includegraphics[height=0.8em]{images/icon-area.png}\textbf{Bones} & \includegraphics[height=0.8em]{images/icon-task.png}Level & \includegraphics[height=0.8em]{images/icon-area.png}Chest & \includegraphics[height=0.8em]{images/icon-task.png}\textbf{Diseases} & \includegraphics[height=0.8em]{images/icon-area.png}\textbf{Bones} & \includegraphics[height=0.8em]{images/icon-task.png}\textbf{Diseases} & 37 & 31 & 43 & \ding{51} \\
\includegraphics[height=0.8em]{images/icon-area.png}\textbf{Bones} & \includegraphics[height=0.8em]{images/icon-task.png}State & \includegraphics[height=0.8em]{images/icon-area.png}Breast & \includegraphics[height=0.8em]{images/icon-task.png}\textbf{Diseases} & \includegraphics[height=0.8em]{images/icon-area.png}\textbf{Bones} & \includegraphics[height=0.8em]{images/icon-task.png}\textbf{Diseases} & 37 & 37 & 43 & \ding{51} \\
\includegraphics[height=0.8em]{images/icon-area.png}\textbf{Bones} & \includegraphics[height=0.8em]{images/icon-task.png}State & \includegraphics[height=0.8em]{images/icon-area.png}Lung & \includegraphics[height=0.8em]{images/icon-task.png}\textbf{Diseases} & \includegraphics[height=0.8em]{images/icon-area.png}\textbf{Bones} & \includegraphics[height=0.8em]{images/icon-task.png}\textbf{Diseases} & 37 & 37 & 43 & \ding{51} \\
\includegraphics[height=0.8em]{images/icon-area.png}\textbf{Bones} & \includegraphics[height=0.8em]{images/icon-task.png}State & \includegraphics[height=0.8em]{images/icon-area.png}Chest & \includegraphics[height=0.8em]{images/icon-task.png}\textbf{Diseases} & \includegraphics[height=0.8em]{images/icon-area.png}\textbf{Bones} & \includegraphics[height=0.8em]{images/icon-task.png}\textbf{Diseases} & 37 & 37 & 41 & \ding{51} \\
\includegraphics[height=0.8em]{images/icon-area.png}\textbf{Lung} & \includegraphics[height=0.8em]{images/icon-task.png}COVID & \includegraphics[height=0.8em]{images/icon-area.png}Breast & \includegraphics[height=0.8em]{images/icon-task.png}\textbf{Diseases} & \includegraphics[height=0.8em]{images/icon-area.png}\textbf{Lung} & \includegraphics[height=0.8em]{images/icon-task.png}\textbf{Diseases} & 49 & 48 & 51 & \ding{51} \\
\includegraphics[height=0.8em]{images/icon-area.png}\textbf{Lung} & \includegraphics[height=0.8em]{images/icon-task.png}COVID & \includegraphics[height=0.8em]{images/icon-area.png}Bones & \includegraphics[height=0.8em]{images/icon-task.png}\textbf{Diseases} & \includegraphics[height=0.8em]{images/icon-area.png}\textbf{Lung} & \includegraphics[height=0.8em]{images/icon-task.png}\textbf{Diseases} & 49 & 48 & 52 & \ding{51} \\
\includegraphics[height=0.8em]{images/icon-area.png}\textbf{Lung} & \includegraphics[height=0.8em]{images/icon-task.png}COVID & \includegraphics[height=0.8em]{images/icon-area.png}Chest & \includegraphics[height=0.8em]{images/icon-task.png}\textbf{Diseases} & \includegraphics[height=0.8em]{images/icon-area.png}\textbf{Lung} & \includegraphics[height=0.8em]{images/icon-task.png}\textbf{Diseases} & 49 & 48 & 51 & \ding{51} \\
\midrule 
 \includegraphics[height=0.8em]{images/icon-modality.png}\textbf{CT} & \includegraphics[height=0.8em]{images/icon-task.png}Cancer & \includegraphics[height=0.8em]{images/icon-modality.png}X-ray & \includegraphics[height=0.8em]{images/icon-task.png}\textbf{COVID} &  \includegraphics[height=0.8em]{images/icon-modality.png}\textbf{CT} & \includegraphics[height=0.8em]{images/icon-task.png}\textbf{COVID} & 47 & 46 & 72 & \ding{51} \\
\includegraphics[height=0.8em]{images/icon-modality.png}\textbf{X-ray} & \includegraphics[height=0.8em]{images/icon-task.png}Diseases & \includegraphics[height=0.8em]{images/icon-modality.png}CT & \includegraphics[height=0.8em]{images/icon-task.png}\textbf{COVID} &  \includegraphics[height=0.8em]{images/icon-modality.png}\textbf{X-ray} & \includegraphics[height=0.8em]{images/icon-task.png}\textbf{COVID} & 30 & 21 & 49 & \ding{51} \\
\includegraphics[height=0.8em]{images/icon-modality.png}\textbf{X-ray} & \includegraphics[height=0.8em]{images/icon-task.png}Diseases & \includegraphics[height=0.8em]{images/icon-modality.png}CT & \includegraphics[height=0.8em]{images/icon-task.png}\textbf{State} &  \includegraphics[height=0.8em]{images/icon-modality.png}\textbf{X-ray} & \includegraphics[height=0.8em]{images/icon-task.png}\textbf{State} & 30 & 21 & 46 & \ding{51} \\
 \includegraphics[height=0.8em]{images/icon-modality.png}\textbf{CT} & \includegraphics[height=0.8em]{images/icon-task.png}State & \includegraphics[height=0.8em]{images/icon-modality.png}X-ray & \includegraphics[height=0.8em]{images/icon-task.png}\textbf{Cancer} &  \includegraphics[height=0.8em]{images/icon-modality.png}\textbf{CT} & \includegraphics[height=0.8em]{images/icon-task.png}\textbf{Cancer} & 33 & 28 & 28 & \ding{55} \\
\midrule 

\includegraphics[height=0.8em]{images/icon-modality.png}\textbf{X-ray} & \includegraphics[height=0.8em]{images/icon-area.png}Bones & \includegraphics[height=0.8em]{images/icon-modality.png}CT & \includegraphics[height=0.8em]{images/icon-area.png}\textbf{Brain} & \includegraphics[height=0.8em]{images/icon-modality.png}\textbf{X-ray} & \includegraphics[height=0.8em]{images/icon-area.png}\textbf{Brain} & 49 & 49 & 91 & \ding{51} \\

\includegraphics[height=0.8em]{images/icon-modality.png}\textbf{X-ray} & \includegraphics[height=0.8em]{images/icon-area.png}Lung & \includegraphics[height=0.8em]{images/icon-modality.png}CT & \includegraphics[height=0.8em]{images/icon-area.png}\textbf{Brain} & \includegraphics[height=0.8em]{images/icon-modality.png}\textbf{X-ray} & \includegraphics[height=0.8em]{images/icon-area.png}\textbf{Brain} & 49 & 50 & 81 & \ding{51} \\

\includegraphics[height=0.8em]{images/icon-modality.png}\textbf{X-ray} & \includegraphics[height=0.8em]{images/icon-area.png}Bones & \includegraphics[height=0.8em]{images/icon-modality.png}CT & \includegraphics[height=0.8em]{images/icon-area.png}\textbf{Brain} & \includegraphics[height=0.8em]{images/icon-modality.png}\textbf{X-ray} & \includegraphics[height=0.8em]{images/icon-area.png}\textbf{Brain} & 25 & 51 & 74 & \ding{51} \\

\includegraphics[height=0.8em]{images/icon-modality.png}\textbf{X-ray} & \includegraphics[height=0.8em]{images/icon-area.png}Lung & \includegraphics[height=0.8em]{images/icon-modality.png}CT & \includegraphics[height=0.8em]{images/icon-area.png}\textbf{Brain} & \includegraphics[height=0.8em]{images/icon-modality.png}\textbf{X-ray} & \includegraphics[height=0.8em]{images/icon-area.png}\textbf{Brain} & 49 & 52 & 52 & \ding{55} \\

\includegraphics[height=0.8em]{images/icon-modality.png}\textbf{CT} & \includegraphics[height=0.8em]{images/icon-area.png}Lung & \includegraphics[height=0.8em]{images/icon-modality.png}X-ray & \includegraphics[height=0.8em]{images/icon-area.png}\textbf{Brain} & \includegraphics[height=0.8em]{images/icon-modality.png}\textbf{CT} & \includegraphics[height=0.8em]{images/icon-area.png}\textbf{Brain} & 33 & 50 & 60 & \ding{51} \\

\includegraphics[height=0.8em]{images/icon-modality.png}\textbf{CT} & \includegraphics[height=0.8em]{images/icon-area.png}Brain & \includegraphics[height=0.8em]{images/icon-modality.png}X-ray & \includegraphics[height=0.8em]{images/icon-area.png}\textbf{Lung} & \includegraphics[height=0.8em]{images/icon-modality.png}\textbf{CT} & \includegraphics[height=0.8em]{images/icon-area.png}\textbf{Lung} & 25 & 25 & 36 & \ding{51} \\

\includegraphics[height=0.8em]{images/icon-modality.png}\textbf{CT} & \includegraphics[height=0.8em]{images/icon-area.png}Brain & \includegraphics[height=0.8em]{images/icon-modality.png}X-ray & \includegraphics[height=0.8em]{images/icon-area.png}\textbf{Lung} & \includegraphics[height=0.8em]{images/icon-modality.png}\textbf{CT} & \includegraphics[height=0.8em]{images/icon-area.png}\textbf{Lung} & 47 & 50 & 81 & \ding{51} \\

\includegraphics[height=0.8em]{images/icon-modality.png}\textbf{CT} & \includegraphics[height=0.8em]{images/icon-area.png}Brain & \includegraphics[height=0.8em]{images/icon-modality.png}X-ray & \includegraphics[height=0.8em]{images/icon-area.png}\textbf{Lung} & \includegraphics[height=0.8em]{images/icon-modality.png}\textbf{CT} & \includegraphics[height=0.8em]{images/icon-area.png}\textbf{Lung} & 47 & 50 & 71 & \ding{51} \\

\includegraphics[height=0.8em]{images/icon-modality.png}\textbf{X-ray} & \includegraphics[height=0.8em]{images/icon-area.png}Bones & \includegraphics[height=0.8em]{images/icon-modality.png}CT & \includegraphics[height=0.8em]{images/icon-area.png}\textbf{Lung} & \includegraphics[height=0.8em]{images/icon-modality.png}\textbf{X-ray} & \includegraphics[height=0.8em]{images/icon-area.png}\textbf{Lung} & 30 & 32 & 28 & \ding{55} \\

\includegraphics[height=0.8em]{images/icon-modality.png}\textbf{X-ray} & \includegraphics[height=0.8em]{images/icon-area.png}Brain & \includegraphics[height=0.8em]{images/icon-modality.png}CT & \includegraphics[height=0.8em]{images/icon-area.png}\textbf{Lung} & \includegraphics[height=0.8em]{images/icon-modality.png}\textbf{X-ray} & \includegraphics[height=0.8em]{images/icon-area.png}\textbf{Lung} & 30 & 32 & 35 & \ding{51} \\

\includegraphics[height=0.8em]{images/icon-modality.png}\textbf{X-ray} & \includegraphics[height=0.8em]{images/icon-area.png}Bones & \includegraphics[height=0.8em]{images/icon-modality.png}CT & \includegraphics[height=0.8em]{images/icon-area.png}\textbf{Lung} & \includegraphics[height=0.8em]{images/icon-modality.png}\textbf{X-ray} & \includegraphics[height=0.8em]{images/icon-area.png}\textbf{Lung} & 30 & 32 & 41 & \ding{51} \\

\includegraphics[height=0.8em]{images/icon-modality.png}\textbf{X-ray} & \includegraphics[height=0.8em]{images/icon-area.png}Brain & \includegraphics[height=0.8em]{images/icon-modality.png}CT & \includegraphics[height=0.8em]{images/icon-area.png}\textbf{Lung} & \includegraphics[height=0.8em]{images/icon-modality.png}\textbf{X-ray} & \includegraphics[height=0.8em]{images/icon-area.png}\textbf{Lung} & 30 & 32 & 42 & \ding{51} \\
\midrule 
\includegraphics[height=0.8em]{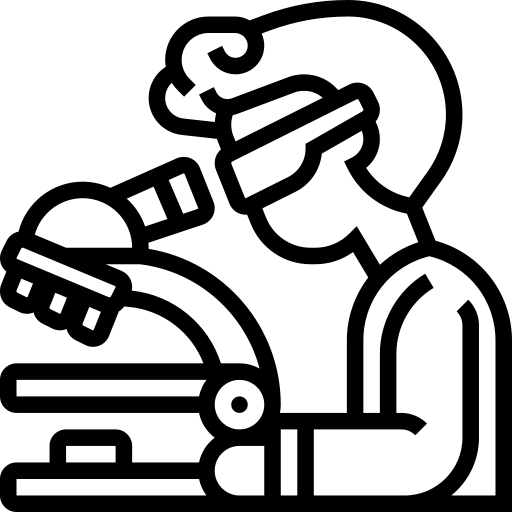}\textbf{Der - Skin} & \includegraphics[height=0.8em]{images/icon-task.png}Cancer & \includegraphics[height=0.8em]{images/icon-modality_and_area.png}FP -  Fundus & \includegraphics[height=0.8em]{images/icon-task.png}\textbf{Diseases} & \includegraphics[height=0.8em]{images/icon-modality_and_area.png}\textbf{Der - Skin} & \includegraphics[height=0.8em]{images/icon-task.png}\textbf{Diseases} & 25 & 29 & 33 & \ding{51} \\
\includegraphics[height=0.8em]{images/icon-modality_and_area.png}\textbf{Der - Skin} & \includegraphics[height=0.8em]{images/icon-task.png}Cancer & \includegraphics[height=0.8em]{images/icon-modality_and_area.png}OCT - Retine & \includegraphics[height=0.8em]{images/icon-task.png}\textbf{Diseases} & \includegraphics[height=0.8em]{images/icon-modality_and_area.png}\textbf{Der - Skin} & \includegraphics[height=0.8em]{images/icon-task.png}\textbf{Diseases} & 25 & 29 & 33 & \ding{51} \\
\includegraphics[height=0.8em]{images/icon-modality_and_area.png}\textbf{Der - Skin} & \includegraphics[height=0.8em]{images/icon-task.png}Diseases & \includegraphics[height=0.8em]{images/icon-modality_and_area.png}DP - Mouth & \includegraphics[height=0.8em]{images/icon-task.png}\textbf{Cancer} & \includegraphics[height=0.8em]{images/icon-modality_and_area.png}\textbf{Der - Skin} & \includegraphics[height=0.8em]{images/icon-task.png}\textbf{Cancer} & 40 & 33 & 63 & \ding{51} \\
\includegraphics[height=0.8em]{images/icon-modality_and_area.png}\textbf{Der - Skin} & \includegraphics[height=0.8em]{images/icon-task.png}Diseases & \includegraphics[height=0.8em]{images/icon-modality_and_area.png}Mic - Cell & \includegraphics[height=0.8em]{images/icon-task.png}\textbf{Cancer} & \includegraphics[height=0.8em]{images/icon-modality_and_area.png}\textbf{Der - Skin} & \includegraphics[height=0.8em]{images/icon-task.png}\textbf{Cancer} & 40 & 33 & 63 & \ding{51} \\
\includegraphics[height=0.8em]{images/icon-modality_and_area.png}\textbf{DP - Mouth} & \includegraphics[height=0.8em]{images/icon-task.png}State & \includegraphics[height=0.8em]{images/icon-modality_and_area.png}Der - Skin & \includegraphics[height=0.8em]{images/icon-task.png}\textbf{Cancer} & \includegraphics[height=0.8em]{images/icon-modality_and_area.png}\textbf{DP - Mouth} & \includegraphics[height=0.8em]{images/icon-task.png}\textbf{Cancer} & 48 & 50 & 52 & \ding{51} \\
\includegraphics[height=0.8em]{images/icon-modality_and_area.png}\textbf{DP - Mouth} & \includegraphics[height=0.8em]{images/icon-task.png}State & \includegraphics[height=0.8em]{images/icon-modality_and_area.png}Mic - Cell & \includegraphics[height=0.8em]{images/icon-task.png}\textbf{Cancer} & \includegraphics[height=0.8em]{images/icon-modality_and_area.png}\textbf{DP - Mouth} & \includegraphics[height=0.8em]{images/icon-task.png}\textbf{Cancer} & 48 & 50 & 55 & \ding{51} \\
\includegraphics[height=0.8em]{images/icon-modality_and_area.png}\textbf{FP - Fundus} & \includegraphics[height=0.8em]{images/icon-task.png}Diseases & \includegraphics[height=0.8em]{images/icon-modality_and_area.png}Mic - Cell & \includegraphics[height=0.8em]{images/icon-task.png}\textbf{Level} & \includegraphics[height=0.8em]{images/icon-modality_and_area.png}\textbf{FP - Fundus} & \includegraphics[height=0.8em]{images/icon-task.png}\textbf{Level} & 33 & 36 & 42 & \ding{51} \\
\includegraphics[height=0.8em]{images/icon-modality_and_area.png}\textbf{Mic - Cell}  & \includegraphics[height=0.8em]{images/icon-task.png}Recognition & \includegraphics[height=0.8em]{images/icon-modality_and_area.png}FP - Fundus & \includegraphics[height=0.8em]{images/icon-task.png}\textbf{Level} & \includegraphics[height=0.8em]{images/icon-modality_and_area.png}\textbf{Mic - Cell} & \includegraphics[height=0.8em]{images/icon-task.png}\textbf{Level} & 23 & 33 & 32 & \ding{55} \\
\includegraphics[height=0.8em]{images/icon-modality_and_area.png}\textbf{Mic - Cell} & \includegraphics[height=0.8em]{images/icon-task.png}Recognition & \includegraphics[height=0.8em]{images/icon-modality_and_area.png}Der - Skin & \includegraphics[height=0.8em]{images/icon-task.png}\textbf{Cancer} & \includegraphics[height=0.8em]{images/icon-modality_and_area.png}\textbf{Mic - Cell} & \includegraphics[height=0.8em]{images/icon-task.png}\textbf{Cancer} & 49 & 50 & 50 & \ding{55} \\
\includegraphics[height=0.8em]{images/icon-modality_and_area.png}\textbf{Mic - Cell} & \includegraphics[height=0.8em]{images/icon-task.png}Recognition & \includegraphics[height=0.8em]{images/icon-modality_and_area.png}DP - Mouth & \includegraphics[height=0.8em]{images/icon-task.png}\textbf{Cancer} & \includegraphics[height=0.8em]{images/icon-modality_and_area.png}\textbf{Mic - Cell} & \includegraphics[height=0.8em]{images/icon-task.png}\textbf{Cancer} & 49 & 51 & 62 & \ding{51} \\
\includegraphics[height=0.8em]{images/icon-modality_and_area.png}\textbf{Mic - Cell} & \includegraphics[height=0.8em]{images/icon-task.png}Level & \includegraphics[height=0.8em]{images/icon-modality_and_area.png}Der - Skin & \includegraphics[height=0.8em]{images/icon-task.png}\textbf{Cancer} & \includegraphics[height=0.8em]{images/icon-modality_and_area.png}\textbf{Mic - Cell} & \includegraphics[height=0.8em]{images/icon-task.png}\textbf{Cancer} & 49 & 51 & 52 & \ding{51} \\
\includegraphics[height=0.8em]{images/icon-modality_and_area.png}\textbf{Mic - Cell} & \includegraphics[height=0.8em]{images/icon-task.png}Level & \includegraphics[height=0.8em]{images/icon-modality_and_area.png}DP - Mouth & \includegraphics[height=0.8em]{images/icon-task.png}\textbf{Cancer} & \includegraphics[height=0.8em]{images/icon-modality_and_area.png}\textbf{Mic - Cell} & \includegraphics[height=0.8em]{images/icon-task.png}\textbf{Cancer} & 49 & 51 & 58 & \ding{51} \\
\includegraphics[height=0.8em]{images/icon-modality_and_area.png}\textbf{Mic - Cell} & \includegraphics[height=0.8em]{images/icon-task.png}Cancer & \includegraphics[height=0.8em]{images/icon-modality_and_area.png}FP - Fundus & \includegraphics[height=0.8em]{images/icon-task.png}\textbf{Level} & \includegraphics[height=0.8em]{images/icon-modality_and_area.png}\textbf{Mic - Cell} & \includegraphics[height=0.8em]{images/icon-task.png}\textbf{Level} & 23 & 24 & 27 & \ding{51} \\
\bottomrule
\end{tabular}
}
\caption{Generalization results on classification datasets: \textit{Related Combination} is the training set, \textit{Target Subset} is the goal. \textit{Baseline}, \textit{Baseline+}, and \textit{Trained} represent the model's accuracy(\%) without training, trained on randomly sampled unrelated data, and trained on related data, respectively. \ding{51} in \textit{CG Helps} indicates successful generalization, while \ding{55} denotes failure. The 4 segmented areas represent different Direction Types: fixed modality \includegraphics[height=0.8em]{images/icon-modality.png}, fixed area \includegraphics[height=0.65em]{images/icon-area.png}, fixed task \includegraphics[height=0.65em]{images/icon-task.png}, and modality-area paired 
combinations \includegraphics[height=0.65em]{images/icon-modality_and_area.png}. Although some combinations share the same name, they differ because they fix different elements.}
\label{tab:Compositional Generalization}
\vspace{-3mm}
\end{table*}

\label{sec:metaCG}

\subsection{Experiment Setup}
To explore the existence of CG from a finer perspective, this section focuses on CG with only two MAT-Triplet elements varying while the third remains constant. Additionally, we identified specific Modality-Area pairs~\includegraphics[height=0.8em]{images/icon-modality_and_area.png}, such as dermoscopy paired consistently with skin, which were treated as a special category. These 4 different fixed formats were classified into distinct \textit{\textbf{Direction Types}}.

We adhered to the training setup described in Section \ref{sec:matsec2} and evaluated the model's performance on the \textit{Target} data. \textit{Baseline} refers to the model without any training, while \textit{Trained} refers to the model trained solely on \textit{Related} data. To ensure that our conclusions are not influenced by the amount of training data, we randomly sampled an equal number of data from the \textit{Unrelated} subsets, and this configuration is referred to as \textit{Baseline+}.

\subsection{Results}

Results are shown in Table \ref{tab:Compositional Generalization} and it can be observed that almost all CG combinations are able to generalize to downstream tasks, highlighting that MLLMs can leverage CG to generalize \textit{Target} data across all Direction Types. Besides that, since this experiment focused solely on two-element tuples, we further investigated three-element tuples in Appendix \ref{sec:ia-3cg}, where we also observed similarly strong generalizations when obtaining MAT-Triplet elements from three different datasets.

\textbf{Take-away 1}: \textit{MLLMs can leverage CG to understand unseen medical images.}

\begin{table*}[ht!]
\setlength{\tabcolsep}{2pt}
\centering
\small
\resizebox{0.88\textwidth}{!}{
\begin{tabular}{ll|ll|ll|rr}
\toprule
\multicolumn{4}{c}{\textbf{Related Combination}} &  \multicolumn{2}{c}{\textbf{Target Subset}} & \textbf{Qwen} & \textbf{Llama} \\
\midrule
\includegraphics[height=0.8em]{images/icon-area.png} \textbf{Bones} & \includegraphics[height=0.8em]{images/icon-task.png} State & \includegraphics[height=0.8em]{images/icon-area.png} Breast & \includegraphics[height=0.8em]{images/icon-task.png} \textbf{Diseases} & \includegraphics[height=0.8em]{images/icon-area.png} \textbf{Bones} & \includegraphics[height=0.8em]{images/icon-task.png} \textbf{Diseases} & $+4$ & $+7$ \\
\includegraphics[height=0.8em]{images/icon-area.png} \textbf{Lung} & \includegraphics[height=0.8em]{images/icon-task.png} COVID & \includegraphics[height=0.8em]{images/icon-area.png} Bones & \includegraphics[height=0.8em]{images/icon-task.png} \textbf{Diseases} & \includegraphics[height=0.8em]{images/icon-area.png} \textbf{Lung} & \includegraphics[height=0.8em]{images/icon-task.png} \textbf{Diseases} & $+11$ & $+11$ \\
\includegraphics[height=0.8em]{images/icon-modality.png} \textbf{X-ray} & \includegraphics[height=0.8em]{images/icon-task.png} Diseases & \includegraphics[height=0.8em]{images/icon-modality.png} CT & \includegraphics[height=0.8em]{images/icon-task.png} \textbf{COVID} & \includegraphics[height=0.8em]{images/icon-modality.png} \textbf{X-ray} & \includegraphics[height=0.8em]{images/icon-task.png} \textbf{COVID} & $+5$ & $+5$ \\
\includegraphics[height=0.8em]{images/icon-modality.png} \textbf{X-ray} & \includegraphics[height=0.8em]{images/icon-task.png} Diseases & \includegraphics[height=0.8em]{images/icon-modality.png} CT & \includegraphics[height=0.8em]{images/icon-task.png} \textbf{State} & \includegraphics[height=0.8em]{images/icon-modality.png} \textbf{X-ray} &  \includegraphics[height=0.8em]{images/icon-task.png} \textbf{State} & $+8$ & $+8$ \\
\includegraphics[height=0.8em]{images/icon-modality.png} \textbf{CT} & \includegraphics[height=0.8em]{images/icon-area.png} Brain & \includegraphics[height=0.8em]{images/icon-modality.png} X-ray & \includegraphics[height=0.8em]{images/icon-area.png} \textbf{Lung} & \includegraphics[height=0.8em]{images/icon-modality.png} \textbf{CT} & \includegraphics[height=0.8em]{images/icon-area.png} \textbf{Lung} & $+1$ & $-2$ \\
\includegraphics[height=0.8em]{images/icon-modality.png} \textbf{CT} & \includegraphics[height=0.8em]{images/icon-area.png} Brain & \includegraphics[height=0.8em]{images/icon-modality.png} X-ray & \includegraphics[height=0.8em]{images/icon-area.png} \textbf{Lung} & \includegraphics[height=0.8em]{images/icon-modality.png} \textbf{CT} & \includegraphics[height=0.8em]{images/icon-area.png} \textbf{Lung} & $+7$ & $+8$ \\
\includegraphics[height=0.8em]{images/icon-modality_and_area.png} \textbf{FP - Fundus} & \includegraphics[height=0.8em]{images/icon-task.png} Diseases & \includegraphics[height=0.8em]{images/icon-modality_and_area.png} Mic - Cell & \includegraphics[height=0.8em]{images/icon-task.png} \textbf{Level} & \includegraphics[height=0.8em]{images/icon-modality_and_area.png} \textbf{FP - Fundus} & \includegraphics[height=0.8em]{images/icon-task.png} \textbf{Level} & $-3$ & $+6$ \\
\includegraphics[height=0.8em]{images/icon-modality_and_area.png} \textbf{Mic - Cell} & \includegraphics[height=0.8em]{images/icon-task.png} Recognition & \includegraphics[height=0.8em]{images/icon-modality_and_area.png} FP - Fundus & \includegraphics[height=0.8em]{images/icon-task.png} \textbf{Level} & \includegraphics[height=0.8em]{images/icon-modality_and_area.png} \textbf{Mic - Cell} & \includegraphics[height=0.8em]{images/icon-task.png} \textbf{Level} & $+7$ & $+22$ \\
\bottomrule
\end{tabular}
}
\caption{Result of Qwen2-VL and Llama-3.2-Vision on selected classification datasets in Med-MAT. \textit{Qwen} and \textit{Llama} represent the accuracy(\%) gains they achieved on the respective backbones through CG.}
\label{tab:extending-cg-to-otherbackbones}
\vspace{-3mm}
\end{table*}

In the \textit{Baseline+} setting, we removed all datasets sharing any MAT-Triplet element with the \textit{Target} data. Consequently, \textit{Baseline+} models perform at near-random levels on the test set, indicating they failed to acquire target-relevant knowledge. This suggests that only datasets related through the MAT-Triplet can help the model learn and generalize to new target tasks.

\textbf{Take-away 2}: \textit{Generalization arises in medical datasets in which at least partial  MAT elements pre-exist during training.}

\subsection{Extending CG to other Backbones}
\label{sec: CG-on-other-backbones}


LLaVA was selected as the baseline because its training data and processes are publicly available, ensuring minimal exposure to medical images and preventing bias in the integration of medical image knowledge into the MLLM. To ensure that the results are not affected by the training data or the visual encoder of LLaVA, we randomly sampled two combinations from each \textit{Direction Type} to investigate CG on Qwen2-VL-7B~\cite{Qwen2VL} and Llama3.2-11B-Vision~\cite{llama3.2}.

Qwen2-VL undergoes additional training on proprietary data based on ViT and incorporates a strategy to adjust the number of vision tokens according to resolution. Llama3.2-Vision, on the other hand, pretrains its own vision encoder from scratch using proprietary data. Thus, both models serve as a means to assess whether MLLMs with different training data and vision encoders can still leverage CG to understand unseen images, ensuring that CG is not merely an artifact of LLaVA’s data fitting or specific to its vision encoder.

Table \ref{tab:extending-cg-to-otherbackbones} presents the experimental results, showing that both selected backbones exhibit a certain degree of generalization across most tasks. This suggests that despite differences in pre-train data and vision encoders, different MLLMs can still leverage CG to understand unseen images.

\textbf{Take-away 3}: \textit{CG persists across different MLLM backbones}.

\begin{figure*}[ht!]
    \centering
    \begin{minipage}[c]{0.48\textwidth}
    \centering
        \includegraphics[width=\textwidth]{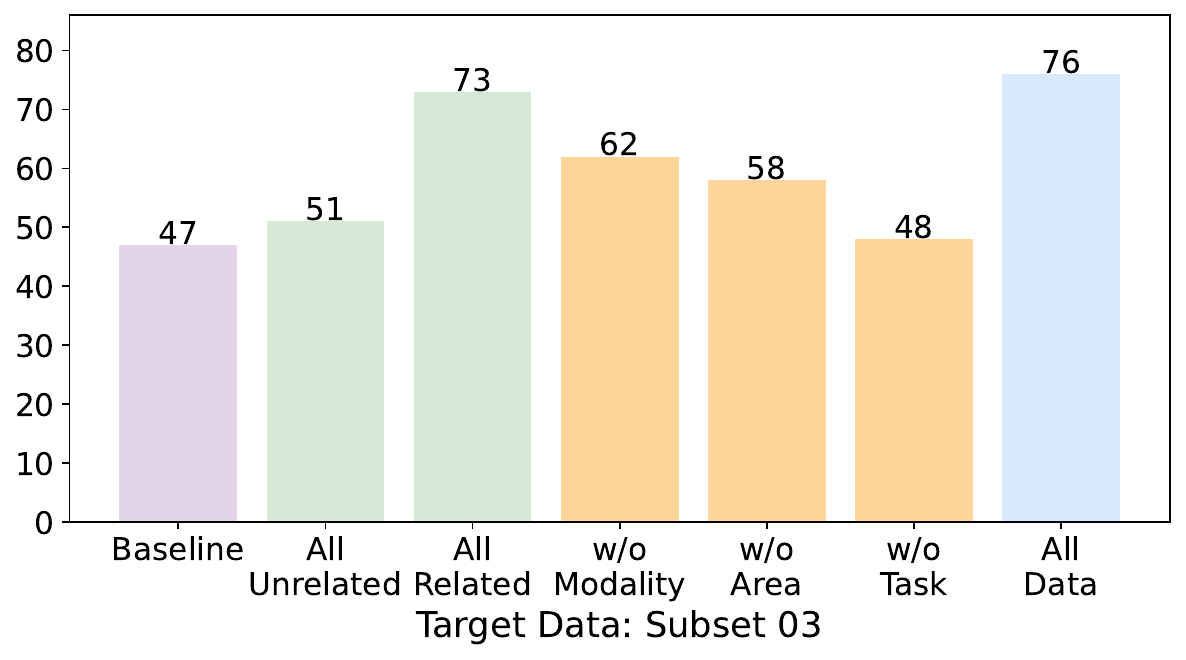}
      \end{minipage}
      \hfill
      \begin{minipage}[c]{0.48\textwidth}
      \centering
        \includegraphics[width=\textwidth]{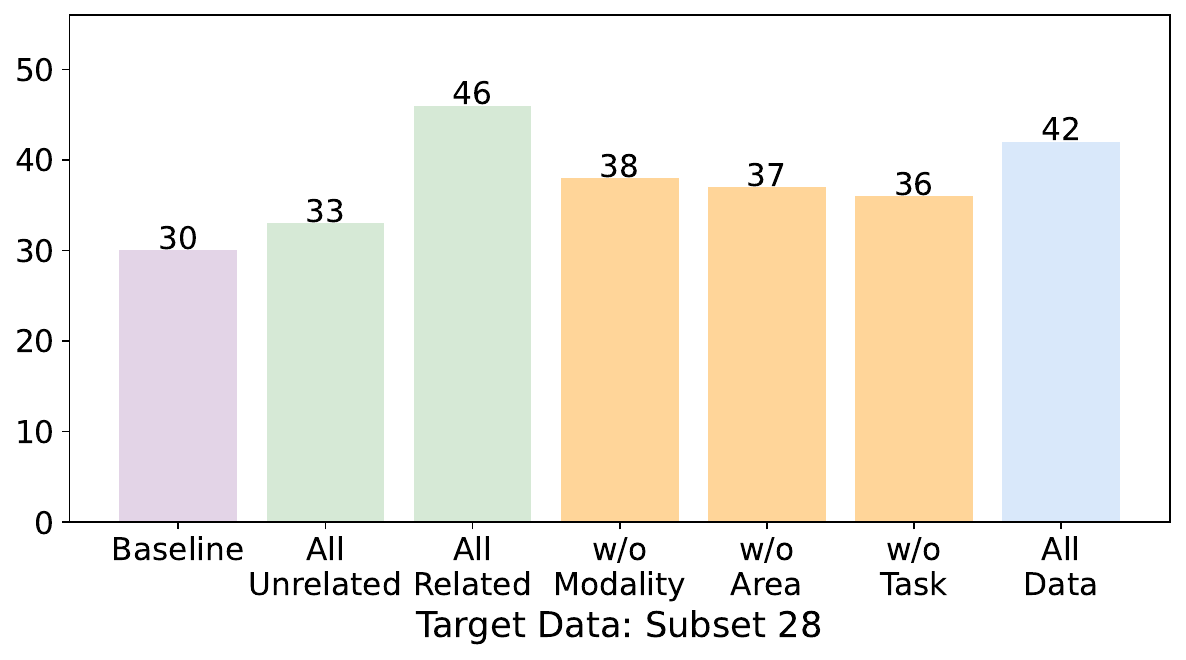}
    \end{minipage}
    \caption{Accuracy(\%) results on the \textit{Target} dataset for various models. \textit{All Related/Unrelated} models are trained on all the related or unrelated datasets of the \textit{Target} Data. \textit{w/o Modality/Area/Task} are trained on All Related datasets but omit those sharing the same element as the \textit{Target} Data, to intentionally disrupt CG. \textit{All Data} uses all available training sets. (Note: The \textit{Target} Data is excluded from training to observe generalization.)}
    \label{fig:Large-scale-composition-generalization}
    \vspace{-5mm}
\end{figure*}

\section{Scaling Combination in CG}
\label{sec:largeCG}

After confirming that CG is indeed a form of generalization in MLLMs, we expanded the number of participating combinations to explore the generalizability of CG and examine its relationship with the generalization exhibited by \textit{Multi-task Training} to address the \textbf{\textit{RQ}}.



\subsection{Experiment Setup}

Two sub-questions have been defined to verify the applicability of CG in multiple data combinations and examine its role in \textit{Multi-task Training}.

\begin{itemize}
    \vspace{-2mm}
    \item \textbf{(Q1)} While previous experiment on CG indicated that \textit{Unrelated} combinations provide no benefit to \textit{Target} data, can generalization arise when training incorporates more \textit{Unrelated} combinations, simulating a multi-task scenario?
    \vspace{-2mm}
    \item \textbf{(Q2)} Previous studies suggest that \textit{Multi-task Training} generally promotes better generalization than single-task training. If the CG conditions in \textit{Multi-task Training} are deliberately disrupted, will the resulting generalization effect be affected?
    \vspace{-2mm}
\end{itemize}

\paragraph{Selection Strategy} To ensure a balanced evaluation of \textit{Related} and \textit{Unrelated} combinations, Subset 03 and Subset 28 were chosen as \textit{Target} datasets because they exhibit the most balanced ratios of \textit{Related} to \textit{Unrelated} subsets (13:11 for Subset 03 and 11:13 for Subset 28), making them ideal for providing a diverse range of compositions in the scale-up experiments.

The baseline was trained on all subsets excluding the \textit{Target} data to evaluate the claim that mixing multi-task data enhances generalization (\colorbox{tableblue!80}{\textit{All Data}}). To construct multiple comparative experiments, models were further trained on either \textit{Related} or \textit{Unrelated} subsets (\colorbox{tablegreen!80}{\textit{All Related} / \textit{All Unrelated}}) to address Q1. For Q2, individual MAT-Triplet elements were systematically removed from the \textit{Related} subsets (\colorbox{tableorange!40}{\textit{Related w/o Modality / Area / Task}}), disrupting CG and assessing the ability to maintain generalization. To ensure consistency, the total data volume in all experiments was limited to 15,000 samples, aligning with the number of ID subsets available after excluding related tasks from Subset 03.

\subsection{Analysis of Scaling Experiment}

Figure \ref{fig:Large-scale-composition-generalization} illustrates the results. It can be observed that even when we expanded the \textit{Unrelated} combination volumes and increased task diversity, the performance of \colorbox{tablegreen!80}{\textit{All Unrelated}} remains close to the \colorbox{tablepurple!80}{\textit{Baseline}}, indicating that these datasets can not support MLLMs to understand the \textit{Target} data.

\textbf{Take-away 4}: \textit{Datasets without MAT-Triplet overlap offer limited benefit for generalization even in the multi-task training scenario (Q1)}.

Besides, \colorbox{tableorange!40}{\textit{w/o Modality / Area / Task}} showed significant accuracy drops compared to \colorbox{tablegreen!80}{\textit{All Related}}, despite holding the training data volume constant. This indicates that if the CG combinations are forcibly disrupted, MLLMs will lose a significant amount of generalization capability for the target data.

\textbf{Take-away 5}: \textit{Disrupting CG leads to a significant decline in generalization ability. (Q2)}.

Notably, \colorbox{tablegreen!80}{\textit{All Related}} achieves a performance level comparable to \colorbox{tableblue!80}{\textit{All Data}}, where all datasets are included in training. This suggests that CG plays a crucial role in enhancing the generalization effect of \textit{Multi-task Training}. Therefore, in conclusion:

\textbf{Take-away 6}: \textit{CG plays an important role in generalization for MLLMs in medical imaging}.


\begin{figure*}[ht!]
    \centering
    \begin{minipage}[c]{0.24\textwidth}
    \centering
    \includegraphics[width=\textwidth]{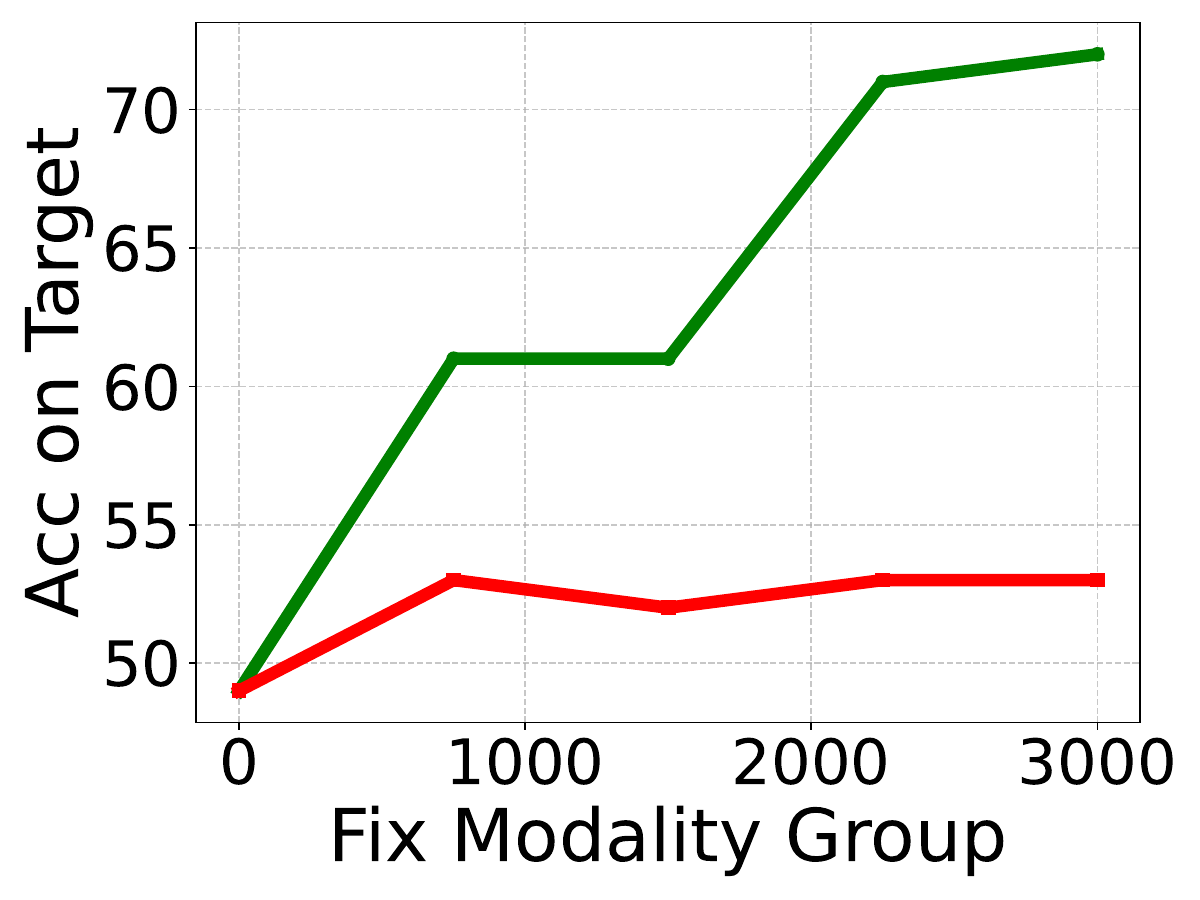}
    \end{minipage}
    \hfill
    \begin{minipage}[c]{0.24\textwidth}
    \centering
    \includegraphics[width=\textwidth]{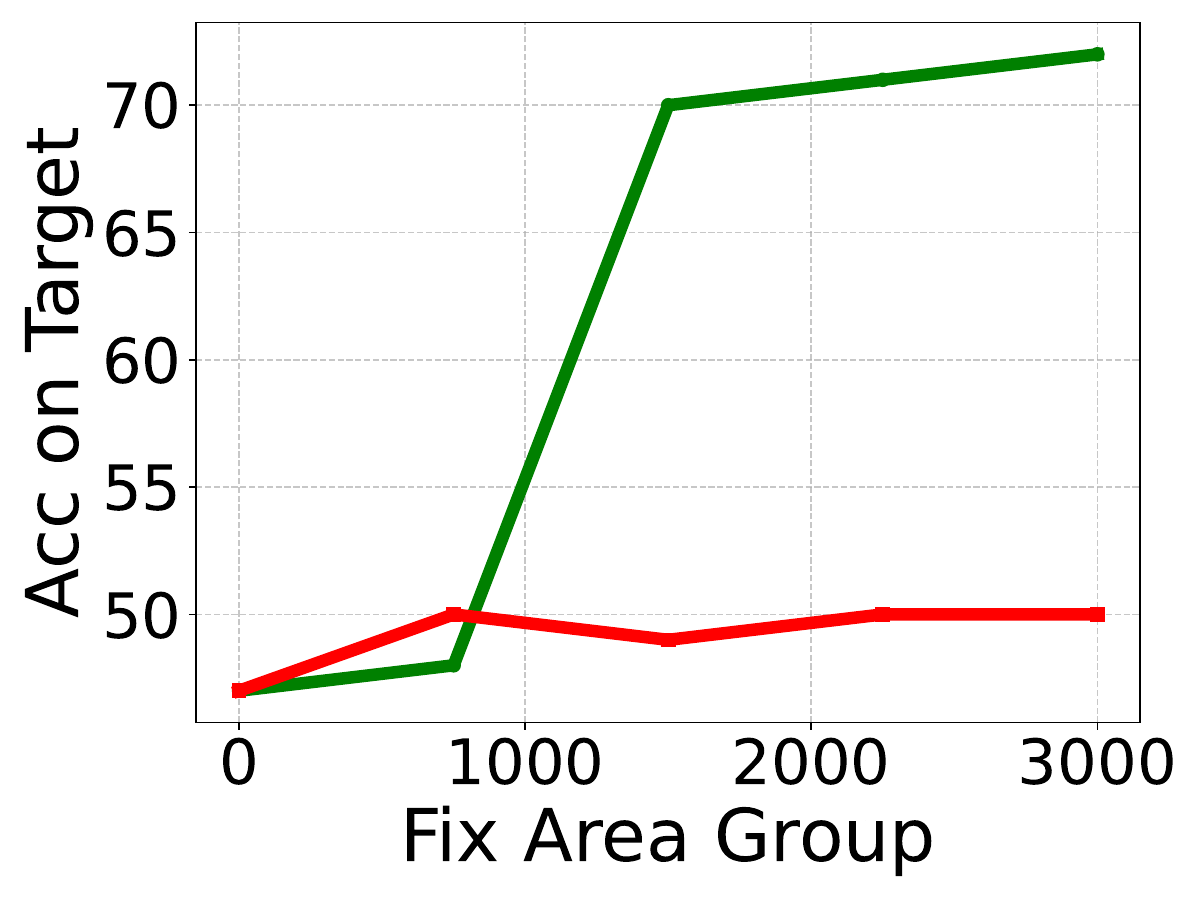}
    \end{minipage}
    \hfill
    \begin{minipage}[c]{0.24\textwidth}
    \centering
    \includegraphics[width=\textwidth]{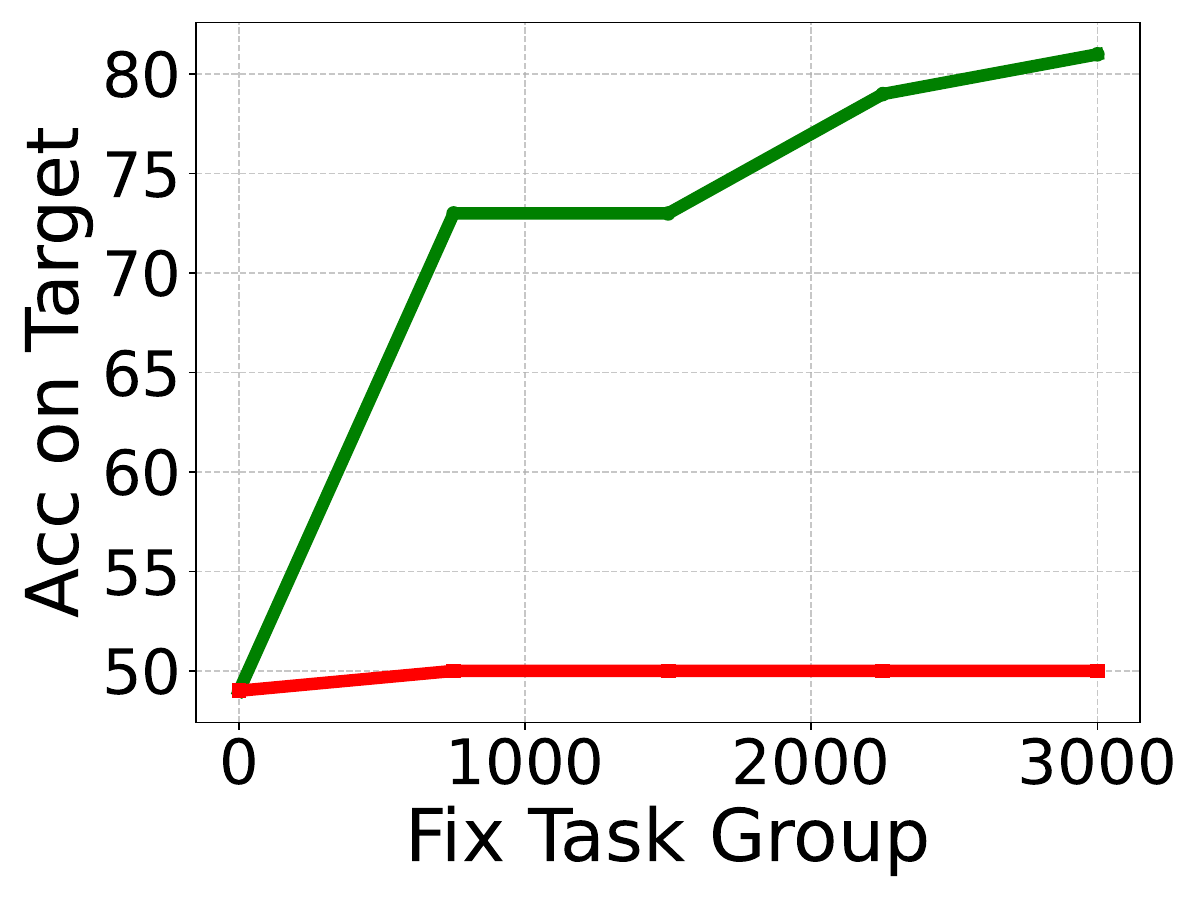}
    \end{minipage}
    \hfill
    \begin{minipage}[c]{0.24\textwidth}
    \centering
    \includegraphics[width=\textwidth]{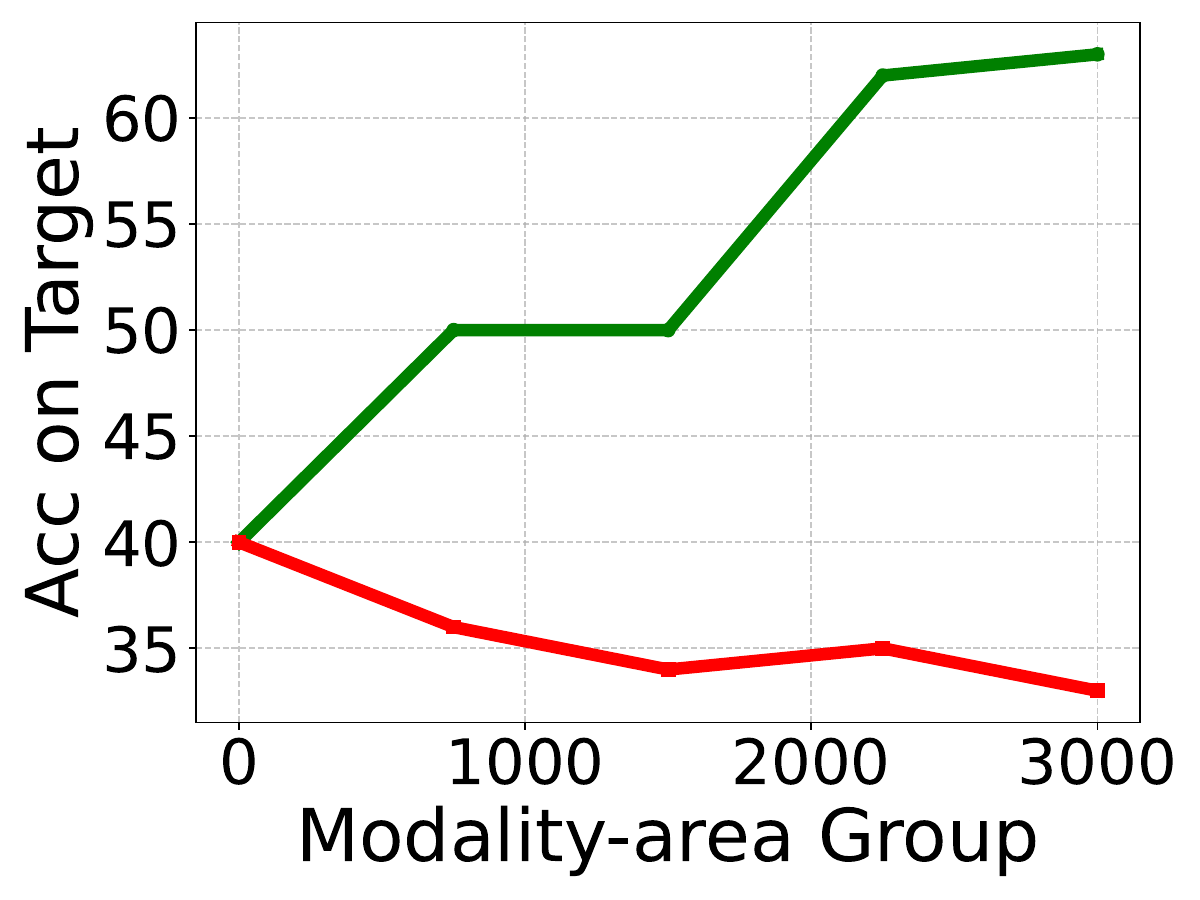}
    \end{minipage}
    \caption{The accuracy curve reflects the impact of gradually increasing the composition dataset size without using \textit{Target} data in training. The green and red lines represent training with \colorbox{customgreen!35}{Related} and \colorbox{customred!35}{Unrelated Data}, respectively.}
    \label{fig:in-depth-1}
    \vspace{-3mm}
\end{figure*}

\begin{figure*}[ht!]
    \centering
    \begin{minipage}[c]{0.24\textwidth}
    \centering
    \includegraphics[width=\textwidth]{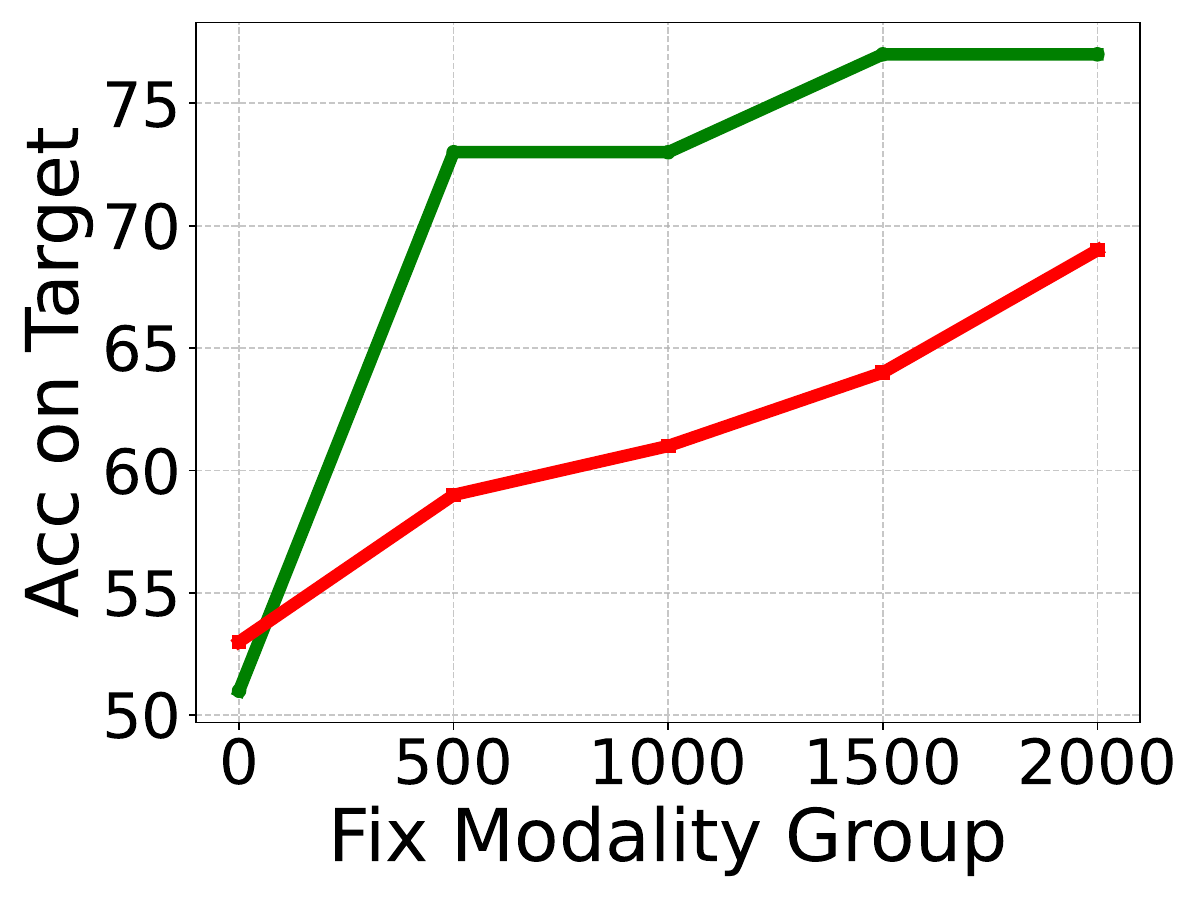}
    \end{minipage}
    \hfill
    \begin{minipage}[c]{0.24\textwidth}
    \centering
    \includegraphics[width=\textwidth]{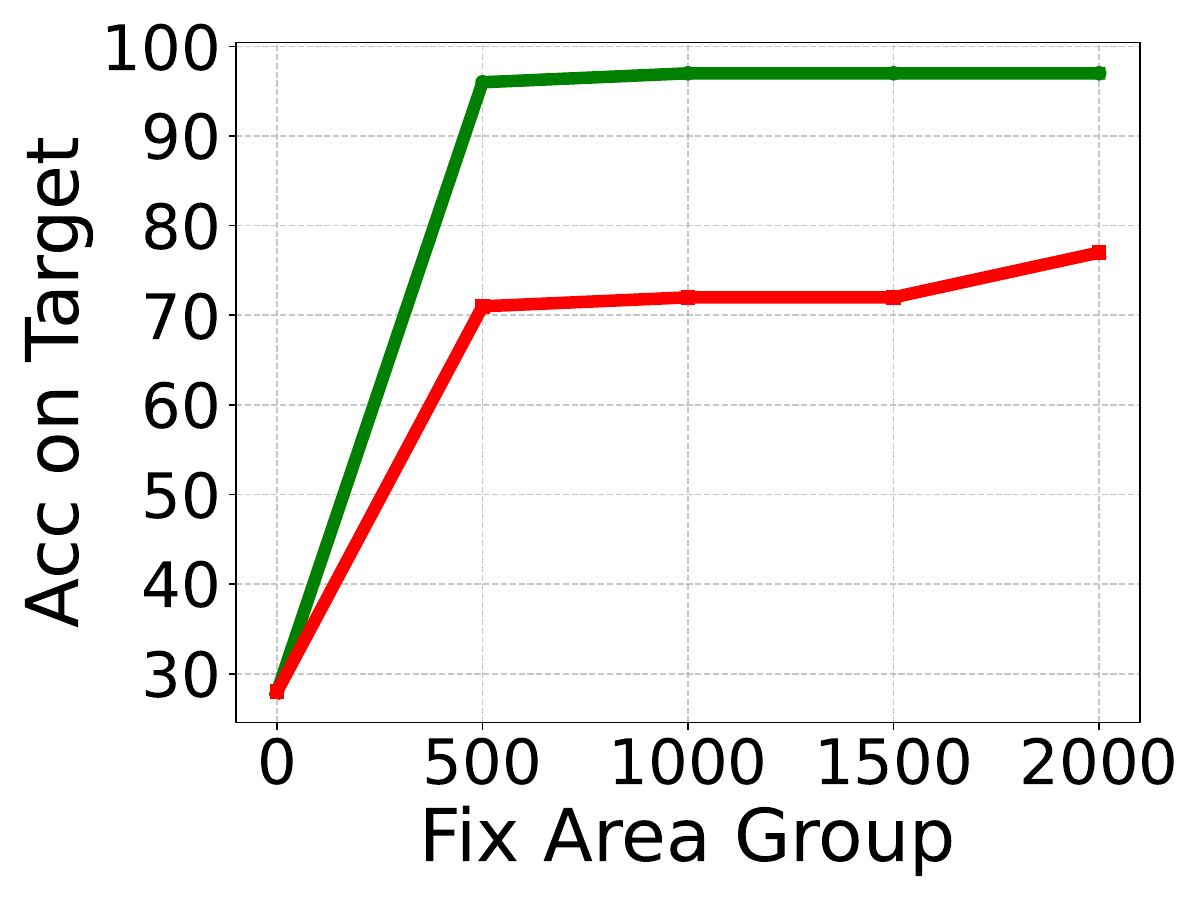}
    \end{minipage}
    \hfill
    \begin{minipage}[c]{0.24\textwidth}
    \centering
    \includegraphics[width=\textwidth]{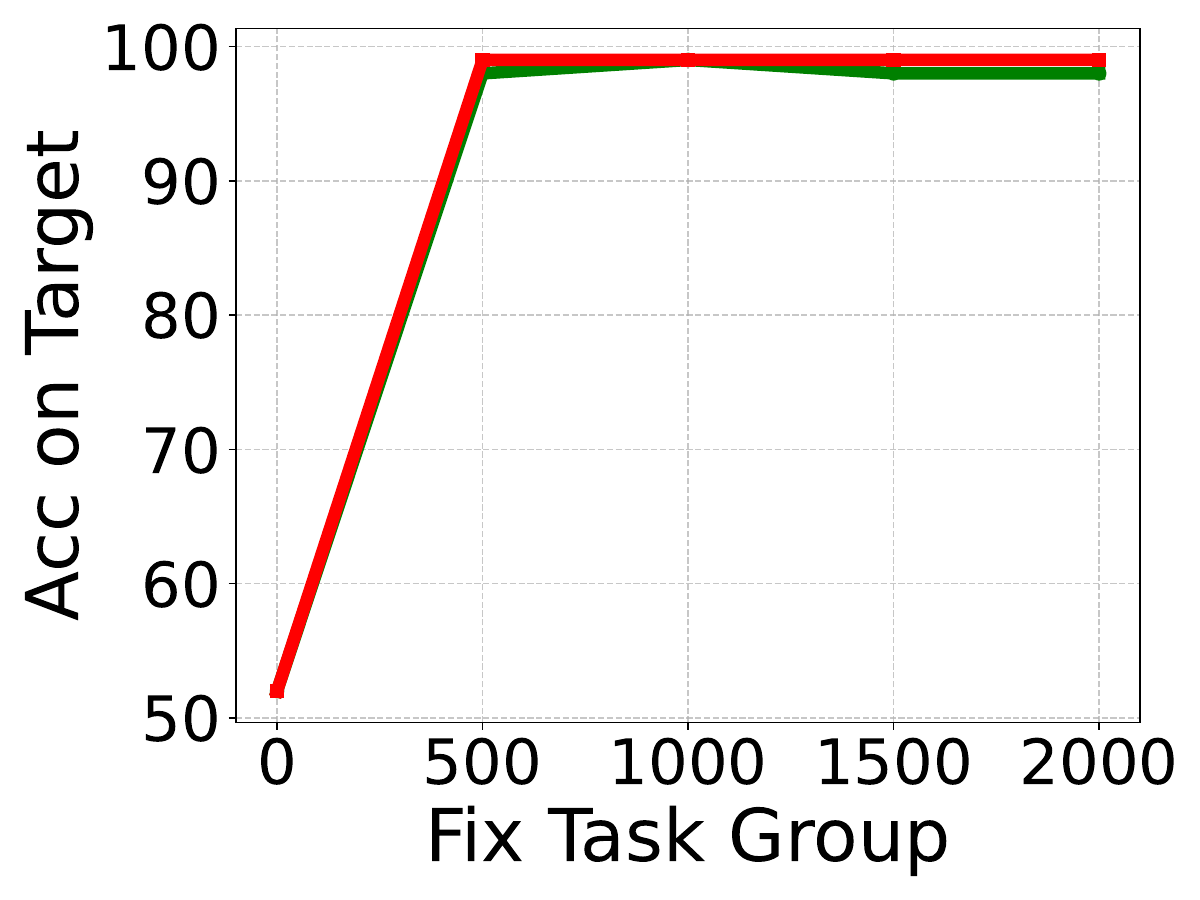}
    \end{minipage}
    \hfill
    \begin{minipage}[c]{0.24\textwidth}
    \centering
    \includegraphics[width=\textwidth]{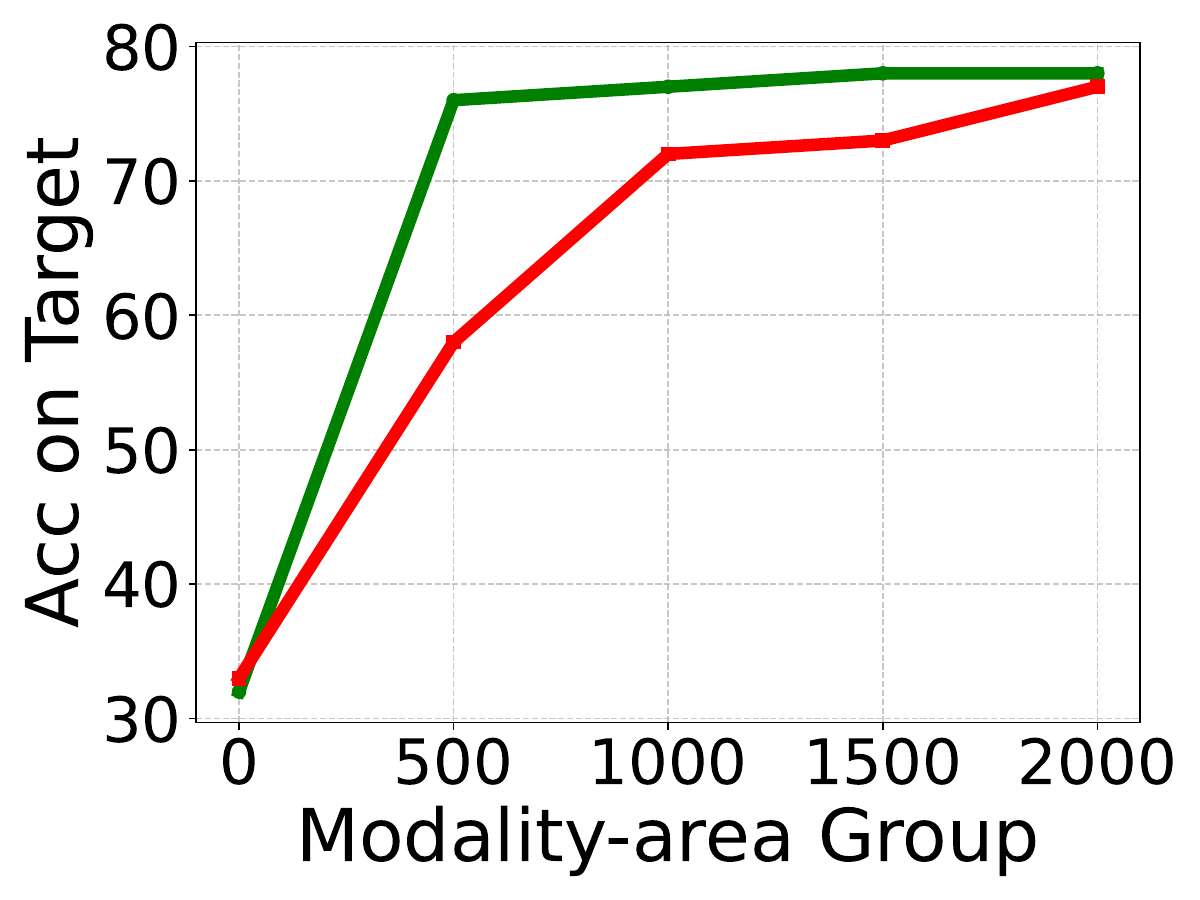}
    \end{minipage}
    \caption{The accuracy curve shows the impact of increasing the composition dataset volume while incorporating \textit{Target} data in training. The green and red lines represent training with \colorbox{customgreen!35}{Related} and \colorbox{customred!35}{Unrelated Data}, respectively.}
    \label{fig:in-depth-2}
    \vspace{-3mm}
\end{figure*}

\section{Potential Applications of CG}


As MLLMs can use CG to generalize unseen medical images, this section attempts to explore its potential applications in training medical MLLMs.

\subsection{Generalization without \textit{Target} Data}
In medical tasks, new and unpredictable conditions, like COVID-19, can emerge at any time. Exploring how to use CG to help MLLMs enhance their ability to identify unknown diseases in the absence of specific datasets is both important and meaningful.

We selected some \textit{Target} datasets and trained the MLLMs using \textit{Related} and \textit{Unrelated} data to observe their generalization to the \textit{Target} data. The generalization trend was assessed by progressively increasing the size of the combination datasets.

\paragraph{Selection Strategy} To highlight the generalization trends, the combinations with strong generalization results were selected from the main experiments. For fairness, we chose the combinations across four types where \textit{Trained} results exceed both \textit{Baseline} and \textit{Baseline}+ by at least 10. If multiple combinations meet the criteria, a random seed of 42 was used to determine the selection.

\paragraph{Analysis} The experimental results are shown in Figure \ref{fig:in-depth-1}, where the red line represents the accuracy curve for \colorbox{customgreen!35}{\textit{Related}} combinations, and the purple line shows the gain from \colorbox{customred!35}{\textit{Unrelated}} combinations. The \textit{Related} combinations group significantly outperformed the \textit{Unrelated} combinations in terms of generalization across all tasks, with this ability continuing to improve as the data size increased. This suggests that \textit{Related} combinations, leveraging CG, enhance the model's ability to understand unknown medical tasks.

\textbf{Take-away 7}: \textit{CG might enable MLLMs to handle tasks without dedicated training data.}


\subsection{Generalization with Limited \textit{Target} Data}
\label{sec:ia-lowdata}

This section investigates the benefit of CG for tasks with limited data, e.g. processing medical images in rare conditions. 


\paragraph{Selection Strategy} To assess generalization in limited data scenarios, we select combinations with poor generalization from Table \ref{tab:Compositional Generalization}. Specifically, for each \textit{Direction Type}, we randomly choose a CG combination with weak generalization (i.e., rows marked with \ding{55} in the last column of Table \ref{tab:Compositional Generalization}). For these combinations, we introduce an additional 2,000 examples from the \textit{Target} data.

\paragraph{Analysis} Figure \ref{fig:in-depth-2} shows the results. It can be seen that as we gradually expand the training volume of \textit{Target} data, adding the \colorbox{customgreen!35}{\textit{Related}} combination for training enabled the model to reach the peak performance more quickly. This suggests that leveraging CG to assist low-data medical scenarios can lead to more data-efficient training, even when CG does not directly result in significant generalization gains in these scenarios.

\textbf{Take-away 8}: \textit{Although CG might not provide direct generalization gains, it helps data efficiency for MLLM training.}



\section{CG across Detection and Classification}
\label{sec: cg-across-location-and-classification}



Previous studies~\cite{ren2024medical, wang2025git} have shown that jointly training classification and detection tasks can mutually enhance their performance. Building on this, we investigate whether MLLMs can leverage classification data (e.g., visual knowledge) and detection data (e.g., spatial information) through CG to improve downstream classification (\textit{\textbf{Q1}}) or detection tasks (\textit{\textbf{Q2}}).

\subsection{Experiment Settings}

\paragraph{Training Setup} Each generalization combination used for training in this experiment includes one detection dataset and one classification dataset to examine the generalization relationship between these two vision tasks. The detailed training parameters can be found in Appendix \ref{app:RQ4}.

\paragraph{Model Selection} Next-Chat~\cite{zhang2023next} and MiniGPT-v2~\cite{chen2023minigpt} are selected as baselines, representing the two main approaches MLLMs use for detection tasks. The former treats bounding boxes as embeddings and decodes them into coordinates using a visual decoder, while the latter processes coordinate points as special text tokens and generates bounding box coordinates directly as output text.

\paragraph{Data Processing} Med-MAT includes both detection and segmentation datasets. If a segmentation dataset provides object localization using masks, we extract the outermost coordinates of the corresponding mask to construct a bounding box, facilitating generalization experiments for detection. Subsequently, to streamline the experiments, we structured the dataset following the official data formats of Next-Chat and MiniGPT-v2.

\begin{figure}[ht!]
    \begin{minipage}[c]{0.23\textwidth}
    \centering
    \includegraphics[width=\textwidth]{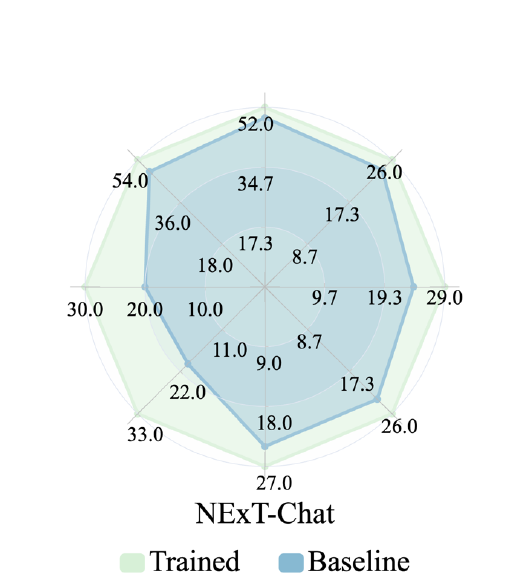}
    \end{minipage}
    \hfill
    \begin{minipage}[c]{0.23\textwidth}
    \centering
    \includegraphics[width=\textwidth]{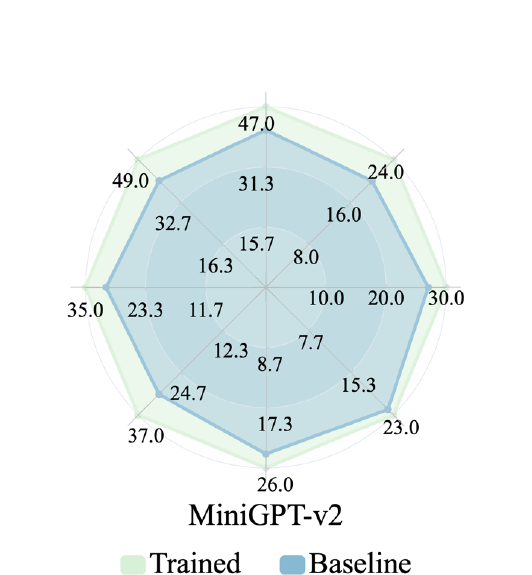}
    \end{minipage}
    \caption{The accuracy(\%) on Classification: \colorbox{customblue!20}{Blue} represents the untrained model, and \colorbox{customgreen!35}{green} represents the CG-trained model. (details in Appendix \ref{app:RQ5})}
    \label{fig:boost-classification}
    \vspace{-5mm}
\end{figure}

\subsection{Benefits for Classification (Q1)}
In this experiment, all possible CG combinations were selected and the CG-trained model will be tested on classification task. The final results in Figure \ref{fig:boost-classification} show that all CG combinations demonstrated the model's successful utilization of detection data for CG to the \textit{Target} data.


\subsection{Benefits for Detection (Q2)}

Subset 38 and 39 are selected as the objects in these datasets are relatively randomly distributed in the images, making them suitable for evaluating the model's detection capability. Subsequently, we selected certain classification datasets to construct CG for testing and used cIoU to evaluate the detection performance (follow \cite{chen2023minigpt}).

Since both baselines lack localization capabilities for medical tasks, we incorporated a fixed amount of \textit{Target} data into our experiments, adjusting the evaluation scenario to assess support in low-data settings. The results in Table \ref{tab:detection-res} show that all selected CG combinations help MLLMs achieve better performance in detection tasks.

\begin{table}[ht!]
\setlength{\tabcolsep}{2pt}
\centering
\small
\resizebox{0.48\textwidth}{!}{
\begin{tabular}{ll|l|rr}
\toprule
\multicolumn{2}{c}{\textbf{Related Combination}} &  \multicolumn{1}{c}{\textbf{Target Subset}} & \textbf{Next-Chat} & \textbf{MiniGPT-v2} \\
\midrule
$D$ - Skin & $C$ - Intestine & $D$ - Intestine & $+3.8$ & $+4.1$ \\
$D$ - Intestine & $C$ - Skin & $D$ - Skin & $+8.4$ & $+7.6$ \\
\bottomrule
\end{tabular}
}
\caption{\textit{Next-Chat} and \textit{MiniGPT-v2} respectively represent the cIoU gain brought by CG. $C$ indicates classification task, $D$ indicates detection task.}
\label{tab:detection-res}
\end{table}

\textbf{Take-away 9}: \textit{MLLMs can perform CG across classification and detection tasks}.

\section{Related Work}
\textbf{Medical MLLMs} Recently, adapting MLLMs to medical tasks has gained prominence due to their success in capturing complex visual features. Current MLLMs typically pair a visual encoder with a text-only LLM, aligning image data with language understanding. Such as Med-Flamingo~\citep{moor2023med} and Med-PaLM~\citep{tu2024towards}, fine-tuned general multimodal models and achieved notable results. Med-Flamingo enhanced OpenFlamingo-9B~\citep{chen2024visual} with medical data, while Med-PaLM adapted PaLM-E~\citep{driess2023palm} using 1 million data points. Similarly, LLaVA-Med~\cite{li2024llava}, Med-Gemini~\citep{saab2024capabilities}, and HuatuoGPT-Vision~\cite{chen2024huatuogpt} utilized specialized datasets and instruction tuning to refine medical VQA tasks.

\paragraph{Generalization on Medical Imaging}  Generalization in medical imaging~\cite{matta2024systematic} has been extensively studied. Early methods utilized data manipulation techniques, such as data augmentation~\cite{li2022single, zhang2022semi}, to enhance model generalization on unseen medical data by adapting to varying distributions. Later approaches focused on representation learning~\cite{le2020contrastive}, preserving essential image information to enable models to handle more complex scenarios. Additionally, some studies~\cite{ren2024medical} explore multiple aspects of medical image processing, examining how classification and segmentation tasks can mutually benefit each other.

\paragraph{Detection with MLLMs} Recent studies employ various strategies to equip MLLMs with the capability to handle detection tasks, such as encoding regions as features to allow models to accept regions as input~\cite{zhang2023gpt4roi}, representing object bounding box coordinates with text tokens~\cite{wang2024visionllm, peng2023kosmos, chen2023shikra}, and employing unique identifiers for task instructions to improve learning efficiency. Additionally, some approaches introduce special tokens to represent images and use their hidden states to decode position information~\cite{zhang2023next, OMGLLaVA}.

\section{Conclusion}

To investigate whether MLLMs can leverage CG to generalize to unseen medical data, we constructed the Med-MAT dataset as a research platform for generalization experiments. The results confirmed the presence of CG and identified it as a key factor of MLLMs' generalization observed in multi-task learning. Further experiments showed that CG helps MLLMs handle limited data conditions, providing support for low-data medical tasks. Additionally, our findings showed that MLLMs can apply CG across detection and classification tasks, underscoring its broad generalization potential.

\section*{Limitations}
The experiment confirms that MLLMs leverage CG for unseen medical images and data-efficient training. However, as shown in Section \ref{sec:largeCG}, disrupting CG reduces generalization but retains some effectiveness, indicating CG is just one aspect of MLLM generalization in medical imaging.


\section*{Potential Risks}
Our research focuses on the compositional generalization of MLLMs on medical images, using data sourced from medical challenges and open-source datasets. However, further experiments are needed to mitigate potential risks when deploying this concept in real-world medical settings.


\section*{Acknowledgments}


This work was supported by Shenzhen Medical Research Fund (No.C2406002) from the Shenzhen Medical Academy of Research and Translation (SMART), the Shenzhen Science and Technology Program (JCYJ20220818103001002), Shenzhen Doctoral Startup Funding (RCBS20221008093330065), Tianyuan Fund for Mathematics of National Natural Science Foundation of China (NSFC) (12326608), Shenzhen Science and Technology Program (Shenzhen Key Laboratory Grant No. ZDSYS20230626091302006), and Shenzhen Stability Science Program 2023.


\bibliography{custom}

\appendix
\section{More Experiments}

\subsection{Benefits for Segmentation}
Segmentation-enabled LLMs, such as Next-GPT, first use the LLM to identify potential regions of the target object and then apply a SAM to decode the object mask, thereby completing the segmentation task. In this context, segmentation can be seen as an extension of detection, potentially requiring more images to achieve improved performance. We conducted additional experiments to explore whether MLLMs can still utilize CG to understand new images across both segmentation and classification tasks.

\begin{table}[ht!]
\setlength{\tabcolsep}{2pt}
\centering
\small
\resizebox{0.38\textwidth}{!}{
\begin{tabular}{ll|l|r}
\toprule
\multicolumn{2}{c}{\textbf{Related Combination}} &  \multicolumn{1}{c}{\textbf{Target Subset}} & \textbf{Next-Chat} \\
\midrule
$D$ - Skin & $C$ - Intestine & $S$ - Intestine & $+7.46$ \\
$D$ - Intestine & $C$ - Skin & $S$ - Skin & $+5.42$ \\
\bottomrule
\end{tabular}
}
\caption{\textit{Next-Chat} represents the cIoU gain brought by CG. $C$ indicates classification task, $S$ indicates Segmentation task.}
\label{tab:segmentation-res}
\vspace{-4mm}
\end{table}

The results in Table~\ref{tab:segmentation-res} demonstrate that, in the context of segmentation tasks, MLLMs are still able to leverage CG to understand new tasks, which is consistent with our original conclusions.

\subsection{More Complex Medical Elements}

While MAT-Triplet Categorization is useful, predefined categories may limit the exploration of more complex medical attributes, so we also considered integrating more flexible categorization to explore additional medical attributes.

\paragraph{Additional Element 1: Population Groups} We selected VinDr-PCXR and MedMAT Subset 31 for the experiment, as they contain X-ray images of children and adult groups, respectively. The results are shown in Table~\ref{tab:moreatt-1}.

\paragraph{Additional Element 2: Finer Disease}

"Finer disease" means more detailed categorization. For instance, we treat COVID and common pneumonia as distinct diseases for generalization. We split the Normal data in the training set into two parts and combined each with COVID and Pneumonia data to create new datasets. The results are shown in Table~\ref{tab:moreatt-2}.

\begin{table}[ht!]
\setlength{\tabcolsep}{2pt}
\centering
\small
\resizebox{0.48\textwidth}{!}{
\begin{tabular}{ll|ll|ll|r}
\toprule
\multicolumn{4}{c}{\textbf{Related Combination}} &  \multicolumn{2}{c}{\textbf{Target Subset}} & \textbf{LLaVA} \\
\midrule
\includegraphics[height=0.8em]{images/icon-modality.png} X-ray & \includegraphics[height=0.8em]{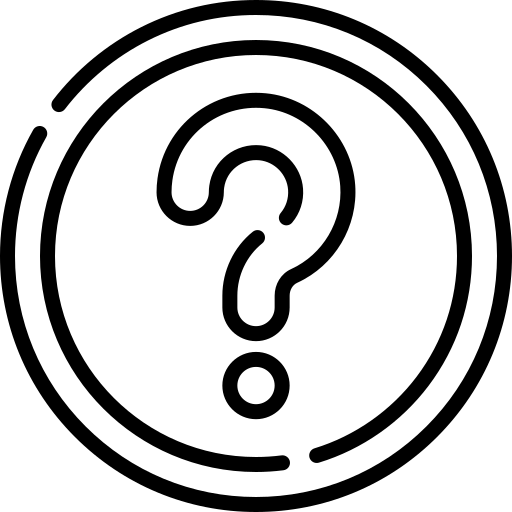} Young & \multicolumn{2}{l|}{\includegraphics[height=0.8em]{images/icon-unknown.png} Unrelated Data} & \includegraphics[height=0.8em]{images/icon-modality.png} X-ray & \includegraphics[height=0.8em]{images/icon-unknown.png} Adults & $+6.04$ \\
\includegraphics[height=0.8em]{images/icon-modality.png} X-ray & \includegraphics[height=0.8em]{images/icon-unknown.png} Young & \includegraphics[height=0.8em]{images/icon-modality.png} CT & Children (CG) & \includegraphics[height=0.8em]{images/icon-modality.png} X-ray & \includegraphics[height=0.8em]{images/icon-unknown.png} Adults & \textbf{$+18.12$} \\
\bottomrule
\end{tabular}
}
\caption{Results of using \textit{Population Groups} as a CG element.}
\label{tab:moreatt-1}
\vspace{-4mm}
\end{table}

\begin{table}[ht!]
\setlength{\tabcolsep}{2pt}
\centering
\small
\resizebox{0.48\textwidth}{!}{
\begin{tabular}{ll|ll|ll|r}
\toprule
\multicolumn{4}{c}{\textbf{Related Combination}} &  \multicolumn{2}{c}{\textbf{Target Subset}} & \textbf{LLaVA} \\
\midrule
\includegraphics[height=0.8em]{images/icon-modality.png} X-ray & \includegraphics[height=0.8em]{images/icon-unknown.png} Pneumonia & \multicolumn{2}{l|}{\includegraphics[height=0.8em]{images/icon-unknown.png} Unrelated Data} & \includegraphics[height=0.8em]{images/icon-modality.png} X-ray & \includegraphics[height=0.8em]{images/icon-unknown.png} COVID & $+11.33$ \\
\includegraphics[height=0.8em]{images/icon-modality.png} X-ray & \includegraphics[height=0.8em]{images/icon-unknown.png} Pneumonia & \includegraphics[height=0.8em]{images/icon-modality.png} CT & \includegraphics[height=0.8em]{images/icon-unknown.png} COVID (CG) & \includegraphics[height=0.8em]{images/icon-modality.png} X-ray & \includegraphics[height=0.8em]{images/icon-unknown.png} COVID & \textbf{$+12.67$} \\
\bottomrule
\end{tabular}
}
\caption{Results of using \textit{Finer Disease} as a CG element.}
\label{tab:moreatt-2}
\vspace{-4mm}
\end{table}

\begin{table*}[ht!]
\setlength{\tabcolsep}{2pt}
\centering
\small
\resizebox{0.95\textwidth}{!}{
\begin{tabular}{ll|ll|ll|r|rrr|r}
\toprule
\multicolumn{4}{c}{\textbf{Related Combination}} &  \multicolumn{2}{c}{\textbf{Target Subset}} & \textbf{Baseline} & \textbf{1st} & \textbf{2nd} & \textbf{3rd} & \textbf{Mean and SD} \\
\midrule
\includegraphics[height=0.8em]{images/icon-area.png} \textbf{Bones} & \includegraphics[height=0.8em]{images/icon-task.png} State & \includegraphics[height=0.8em]{images/icon-area.png} Breast & \includegraphics[height=0.8em]{images/icon-task.png} \textbf{Diseases} & \includegraphics[height=0.8em]{images/icon-area.png} \textbf{Bones} & \includegraphics[height=0.8em]{images/icon-task.png} \textbf{Diseases} & $37.31$ & $43.28$ & $44.78$ & $43.28$ & $43.78 \pm 0.87$ \\
\includegraphics[height=0.8em]{images/icon-area.png} \textbf{Lung} & \includegraphics[height=0.8em]{images/icon-task.png} COVID & \includegraphics[height=0.8em]{images/icon-area.png} Bones & \includegraphics[height=0.8em]{images/icon-task.png} \textbf{Diseases} & \includegraphics[height=0.8em]{images/icon-area.png} \textbf{Lung} & \includegraphics[height=0.8em]{images/icon-task.png} \textbf{Diseases} & $49.00$ & $52.00$ & $52.00$ & $52.00$ & $52.00 \pm 0.00$ \\
\includegraphics[height=0.8em]{images/icon-modality.png} \textbf{X-ray} & \includegraphics[height=0.8em]{images/icon-task.png} Diseases & \includegraphics[height=0.8em]{images/icon-modality.png} CT & \includegraphics[height=0.8em]{images/icon-task.png} \textbf{COVID} & \includegraphics[height=0.8em]{images/icon-modality.png} \textbf{X-ray} & \includegraphics[height=0.8em]{images/icon-task.png} \textbf{COVID} & $30.00$ & $47.33$ & $49.33$ & $49.33$ & $48.66 \pm 1.15$ \\
\includegraphics[height=0.8em]{images/icon-modality.png} \textbf{X-ray} & \includegraphics[height=0.8em]{images/icon-task.png} Diseases & \includegraphics[height=0.8em]{images/icon-modality.png} CT & \includegraphics[height=0.8em]{images/icon-task.png} \textbf{State} & \includegraphics[height=0.8em]{images/icon-modality.png} \textbf{X-ray} &  \includegraphics[height=0.8em]{images/icon-task.png} \textbf{State} & $30.00$ & $46.00$ & $45.33$ & $44.67$ & $45.33 \pm 0.67$ \\
\includegraphics[height=0.8em]{images/icon-modality.png} \textbf{CT} & \includegraphics[height=0.8em]{images/icon-area.png} Brain & \includegraphics[height=0.8em]{images/icon-modality.png} X-ray & \includegraphics[height=0.8em]{images/icon-area.png} \textbf{Lung} & \includegraphics[height=0.8em]{images/icon-modality.png} \textbf{CT} & \includegraphics[height=0.8em]{images/icon-area.png} \textbf{Lung} & $25.00$ & $31.50$ & $32.00$ & $32.00$ & $31.83 \pm 0.29$ \\
\includegraphics[height=0.8em]{images/icon-modality.png} \textbf{CT} & \includegraphics[height=0.8em]{images/icon-area.png} Brain & \includegraphics[height=0.8em]{images/icon-modality.png} X-ray & \includegraphics[height=0.8em]{images/icon-area.png} \textbf{Lung} & \includegraphics[height=0.8em]{images/icon-modality.png} \textbf{CT} & \includegraphics[height=0.8em]{images/icon-area.png} \textbf{Lung} & $47.00$ & $71.00$ & $71.00$ & $70.00$ & $70.67 \pm 0.58$ \\
\includegraphics[height=0.8em]{images/icon-modality_and_area.png} \textbf{FP - Fundus} & \includegraphics[height=0.8em]{images/icon-task.png} Diseases & \includegraphics[height=0.8em]{images/icon-modality_and_area.png} Mic - Cell & \includegraphics[height=0.8em]{images/icon-task.png} \textbf{Level} & \includegraphics[height=0.8em]{images/icon-modality_and_area.png} \textbf{FP - Fundus} & \includegraphics[height=0.8em]{images/icon-task.png} \textbf{Level} & $33.33$ & $42.42$ & $45.45$ & $45.45$ & $44.44 \pm 1.75$ \\
\includegraphics[height=0.8em]{images/icon-modality_and_area.png} \textbf{Mic - Cell} & \includegraphics[height=0.8em]{images/icon-task.png} Recognition & \includegraphics[height=0.8em]{images/icon-modality_and_area.png} FP - Fundus & \includegraphics[height=0.8em]{images/icon-task.png} \textbf{Level} & \includegraphics[height=0.8em]{images/icon-modality_and_area.png} \textbf{Mic - Cell} & \includegraphics[height=0.8em]{images/icon-task.png} \textbf{Level} & $23.00$ & $32.00$ & $32.00$ & $31.50$ & $31.83 \pm 0.29$ \\
\bottomrule
\end{tabular}
}
\caption{Statistical tests of CG experiments. The 1st, 2nd, and 3rd show the generalization results of the experiment in different runs. "Mean" and "SD" represent the average accuracy (\%) and standard deviation.}
\label{tab:statistical-tests}
\end{table*}

\subsection{Statistical Tests of the Generalization Results}

To ensure consistency and repeatability of the experiment, we performed statistical tests in this section. LLaVA is selected as the baseline, and we used the same data combinations from Section~\ref{sec: CG-on-other-backbones}. Each experiment was repeated 3 times, and we reported the mean and standard deviation (SD) of the results.

From the results in Table~\ref{tab:statistical-tests}, we can observe that the outcomes across runs show low variance, indicating overall stability, and they continue to support our original experimental conclusions.

\subsection{CG with All MAT-Triplet Elements from Different Sources}
\label{sec:ia-3cg}

In previous controlled experiments (Section \ref{sec:metaCG}), one element of the MAT-Triplet was kept constant while CG was explored in the remaining two elements. To ensure that all the 3 MAT-Triplet elements of the target data originated from three distinct datasets, additional experiments were conducted to further validate the effectiveness of CG. For these experiments, all possible combinations meeting the criteria in Med-MAT were selected (\textbf{Selection Strategy}). The results presented in Table \ref{tab:CG-3trainingsets} demonstrate that most combinations can effectively generalize to the \textit{Target} data.

\begin{table}[ht!]
\setlength{\tabcolsep}{2pt}
\centering
\small
\resizebox{0.48\textwidth}{!}{
\begin{tabular}{lll|lll|ccc}
\toprule
\multicolumn{3}{c}{\textbf{Related Combination}} &  \multicolumn{3}{c}{\textbf{Target Subset}} & \textbf{Baseline} & \textbf{Trained} & \textbf{CG Helps}\\
\midrule
\includegraphics[height=0.8em]{images/icon-modality.png} CT & \includegraphics[height=0.8em]{images/icon-area.png} Brain & \includegraphics[height=0.8em]{images/icon-task.png} Cancer & \includegraphics[height=0.8em]{images/icon-modality.png} CT & \includegraphics[height=0.8em]{images/icon-area.png} Brain & \includegraphics[height=0.8em]{images/icon-task.png} Cancer & 28 & 26 & \ding{55} \\
\includegraphics[height=0.8em]{images/icon-modality.png} CT & \includegraphics[height=0.8em]{images/icon-area.png} Brain & \includegraphics[height=0.8em]{images/icon-task.png} Cancer & \includegraphics[height=0.8em]{images/icon-modality.png} CT & \includegraphics[height=0.8em]{images/icon-area.png} Brain & \includegraphics[height=0.8em]{images/icon-task.png} Cancer & 28 & 25 & \ding{55} \\
\includegraphics[height=0.8em]{images/icon-modality.png} CT & \includegraphics[height=0.8em]{images/icon-area.png} Brain & \includegraphics[height=0.8em]{images/icon-task.png} State & \includegraphics[height=0.8em]{images/icon-modality.png} CT & \includegraphics[height=0.8em]{images/icon-area.png} Brain & \includegraphics[height=0.8em]{images/icon-task.png} State & 33 & 64 & \ding{51} \\
\includegraphics[height=0.8em]{images/icon-modality.png} CT & \includegraphics[height=0.8em]{images/icon-area.png} Brain & \includegraphics[height=0.8em]{images/icon-task.png} State & \includegraphics[height=0.8em]{images/icon-modality.png} CT & \includegraphics[height=0.8em]{images/icon-area.png} Brain & \includegraphics[height=0.8em]{images/icon-task.png} State & 33 & 70 & \ding{51} \\
\includegraphics[height=0.8em]{images/icon-modality.png} X-ray & \includegraphics[height=0.8em]{images/icon-area.png} Lung & \includegraphics[height=0.8em]{images/icon-task.png} Diseases & \includegraphics[height=0.8em]{images/icon-modality.png} X-ray & \includegraphics[height=0.8em]{images/icon-area.png} Lung & \includegraphics[height=0.8em]{images/icon-task.png} Diseases & 30 & 45 & \ding{51} \\
\includegraphics[height=0.8em]{images/icon-modality.png} X-ray & \includegraphics[height=0.8em]{images/icon-area.png} Lung & \includegraphics[height=0.8em]{images/icon-task.png} Diseases & \includegraphics[height=0.8em]{images/icon-modality.png} X-ray & \includegraphics[height=0.8em]{images/icon-area.png} Lung & \includegraphics[height=0.8em]{images/icon-task.png} Diseases & 30 & 38 & \ding{51} \\
\includegraphics[height=0.8em]{images/icon-modality.png} X-ray & \includegraphics[height=0.8em]{images/icon-area.png} Lung & \includegraphics[height=0.8em]{images/icon-task.png} Diseases & \includegraphics[height=0.8em]{images/icon-modality.png} X-ray & \includegraphics[height=0.8em]{images/icon-area.png} Lung & \includegraphics[height=0.8em]{images/icon-task.png} Diseases & 30 & 44 & \ding{51} \\
\includegraphics[height=0.8em]{images/icon-modality.png} X-ray & \includegraphics[height=0.8em]{images/icon-area.png} Breast & \includegraphics[height=0.8em]{images/icon-task.png} Diseases & \includegraphics[height=0.8em]{images/icon-modality.png} X-ray & \includegraphics[height=0.8em]{images/icon-area.png} Breast & \includegraphics[height=0.8em]{images/icon-task.png} Diseases & 31 & 32 & \ding{51} \\
\includegraphics[height=0.8em]{images/icon-modality.png} X-ray & \includegraphics[height=0.8em]{images/icon-area.png} Breast & \includegraphics[height=0.8em]{images/icon-task.png} Diseases & \includegraphics[height=0.8em]{images/icon-modality.png} X-ray & \includegraphics[height=0.8em]{images/icon-area.png} Breast & \includegraphics[height=0.8em]{images/icon-task.png} Diseases & 31 & 52 & \ding{51} \\
\bottomrule
\end{tabular}
}
\caption{Results from 3 datasets providing different elements of MAT-Triplet. \ding{51} in \textit{CG Helps} indicates successful generalization, while \ding{55} denotes failure.}
\label{tab:CG-3trainingsets}
\vspace{-4mm}
\end{table}

\begin{table*}[ht!]
\setlength{\tabcolsep}{2pt}
\centering
\small
\resizebox{0.87\textwidth}{!}{
\begin{tabular}{ll|ll|ll|ccc}
\toprule
\multicolumn{4}{c}{\textbf{Related Combination}} &  \multicolumn{2}{c}{\textbf{Target Subset}} & \textbf{Baseline} & \textbf{Trained} & \textbf{CG Helps} \\
\midrule
\includegraphics[height=0.8em]{images/icon-area.png} \textbf{Bones} & \includegraphics[height=0.8em]{images/icon-task.png} State & \includegraphics[height=0.8em]{images/icon-area.png} Breast & \includegraphics[height=0.8em]{images/icon-task.png} \textbf{Diseases} & \includegraphics[height=0.8em]{images/icon-area.png} \textbf{Bones} & \includegraphics[height=0.8em]{images/icon-task.png} \textbf{Diseases} & 61 & 65 & \ding{51} \\
\includegraphics[height=0.8em]{images/icon-area.png} \textbf{Lung} & \includegraphics[height=0.8em]{images/icon-task.png} COVID & \includegraphics[height=0.8em]{images/icon-area.png} Bones & \includegraphics[height=0.8em]{images/icon-task.png} \textbf{Diseases} & \includegraphics[height=0.8em]{images/icon-area.png} \textbf{Lung} & \includegraphics[height=0.8em]{images/icon-task.png} \textbf{Diseases} & 80 & 91 & \ding{51} \\
\includegraphics[height=0.8em]{images/icon-modality.png} \textbf{X-ray} & \includegraphics[height=0.8em]{images/icon-task.png} Diseases & \includegraphics[height=0.8em]{images/icon-modality.png} CT & \includegraphics[height=0.8em]{images/icon-task.png} \textbf{COVID} & \includegraphics[height=0.8em]{images/icon-modality.png} \textbf{X-ray} & \includegraphics[height=0.8em]{images/icon-task.png} \textbf{COVID} & 35 & 40 & \ding{51} \\
\includegraphics[height=0.8em]{images/icon-modality.png} \textbf{X-ray} & \includegraphics[height=0.8em]{images/icon-task.png} Diseases & \includegraphics[height=0.8em]{images/icon-modality.png} CT & \includegraphics[height=0.8em]{images/icon-task.png} \textbf{State} & \includegraphics[height=0.8em]{images/icon-modality.png} \textbf{X-ray} &  \includegraphics[height=0.8em]{images/icon-task.png} \textbf{State} & 35 & 43 & \ding{51} \\
\includegraphics[height=0.8em]{images/icon-modality.png} \textbf{CT} & \includegraphics[height=0.8em]{images/icon-area.png} Brain & \includegraphics[height=0.8em]{images/icon-modality.png} X-ray & \includegraphics[height=0.8em]{images/icon-area.png} \textbf{Lung} & \includegraphics[height=0.8em]{images/icon-modality.png} \textbf{CT} & \includegraphics[height=0.8em]{images/icon-area.png} \textbf{Lung} & 32 & 33 & \ding{51} \\
\includegraphics[height=0.8em]{images/icon-modality.png} \textbf{CT} & \includegraphics[height=0.8em]{images/icon-area.png} Brain & \includegraphics[height=0.8em]{images/icon-modality.png} X-ray & \includegraphics[height=0.8em]{images/icon-area.png} \textbf{Lung} & \includegraphics[height=0.8em]{images/icon-modality.png} \textbf{CT} & \includegraphics[height=0.8em]{images/icon-area.png} \textbf{Lung} & 65 & 72 & \ding{51} \\
\includegraphics[height=0.8em]{images/icon-modality_and_area.png} \textbf{FP - Fundus} & \includegraphics[height=0.8em]{images/icon-task.png} Diseases & \includegraphics[height=0.8em]{images/icon-modality_and_area.png} Mic - Cell & \includegraphics[height=0.8em]{images/icon-task.png} \textbf{Level} & \includegraphics[height=0.8em]{images/icon-modality_and_area.png} \textbf{FP - Fundus} & \includegraphics[height=0.8em]{images/icon-task.png} \textbf{Level} & 48 & 45 & \ding{55} \\
\includegraphics[height=0.8em]{images/icon-modality_and_area.png} \textbf{Mic - Cell} & \includegraphics[height=0.8em]{images/icon-task.png} Recognition & \includegraphics[height=0.8em]{images/icon-modality_and_area.png} FP - Fundus & \includegraphics[height=0.8em]{images/icon-task.png} \textbf{Level} & \includegraphics[height=0.8em]{images/icon-modality_and_area.png} \textbf{Mic - Cell} & \includegraphics[height=0.8em]{images/icon-task.png} \textbf{Level} & 34 & 41 & \ding{51} \\
\bottomrule
\end{tabular}
}
\caption{Result of Qwen2-VL on selected classification datasets in Med-MAT. \ding{51} in \textit{CG Helps} indicates successful generalization, while \ding{55} denotes failure.}
\label{tab:qwen2-vl-7b-exp}
\end{table*}

\begin{table*}[ht!]
\setlength{\tabcolsep}{2pt}
\centering
\small
\resizebox{0.87\textwidth}{!}{
\begin{tabular}{ll|ll|ll|ccc}
\toprule
\multicolumn{4}{c}{\textbf{Related Combination}} &  \multicolumn{2}{c}{\textbf{Target Subset}} & \textbf{Baseline} & \textbf{Trained} & \textbf{CG Helps} \\
\midrule
\includegraphics[height=0.8em]{images/icon-area.png} \textbf{Bones} & \includegraphics[height=0.8em]{images/icon-task.png} State & \includegraphics[height=0.8em]{images/icon-area.png} Breast & \includegraphics[height=0.8em]{images/icon-task.png} \textbf{Diseases} & \includegraphics[height=0.8em]{images/icon-area.png} \textbf{Bones} & \includegraphics[height=0.8em]{images/icon-task.png} \textbf{Diseases} & 52 & 59 & \ding{51} \\
\includegraphics[height=0.8em]{images/icon-area.png} \textbf{Lung} & \includegraphics[height=0.8em]{images/icon-task.png} COVID & \includegraphics[height=0.8em]{images/icon-area.png} Bones & \includegraphics[height=0.8em]{images/icon-task.png} \textbf{Diseases} & \includegraphics[height=0.8em]{images/icon-area.png} \textbf{Lung} & \includegraphics[height=0.8em]{images/icon-task.png} \textbf{Diseases} & 64 & 75 & \ding{51} \\
\includegraphics[height=0.8em]{images/icon-modality.png} \textbf{X-ray} & \includegraphics[height=0.8em]{images/icon-task.png} Diseases & \includegraphics[height=0.8em]{images/icon-modality.png} CT & \includegraphics[height=0.8em]{images/icon-task.png} \textbf{COVID} & \includegraphics[height=0.8em]{images/icon-modality.png} \textbf{X-ray} & \includegraphics[height=0.8em]{images/icon-task.png} \textbf{COVID} & 33 & 38 & \ding{51} \\
\includegraphics[height=0.8em]{images/icon-modality.png} \textbf{X-ray} & \includegraphics[height=0.8em]{images/icon-task.png} Diseases & \includegraphics[height=0.8em]{images/icon-modality.png} CT & \includegraphics[height=0.8em]{images/icon-task.png} \textbf{State} & \includegraphics[height=0.8em]{images/icon-modality.png} \textbf{X-ray} &  \includegraphics[height=0.8em]{images/icon-task.png} \textbf{State} & 33 & 41 & \ding{51} \\
\includegraphics[height=0.8em]{images/icon-modality.png} \textbf{CT} & \includegraphics[height=0.8em]{images/icon-area.png} Brain & \includegraphics[height=0.8em]{images/icon-modality.png} X-ray & \includegraphics[height=0.8em]{images/icon-area.png} \textbf{Lung} & \includegraphics[height=0.8em]{images/icon-modality.png} \textbf{CT} & \includegraphics[height=0.8em]{images/icon-area.png} \textbf{Lung} & 31 & 29 & \ding{55} \\
\includegraphics[height=0.8em]{images/icon-modality.png} \textbf{CT} & \includegraphics[height=0.8em]{images/icon-area.png} Brain & \includegraphics[height=0.8em]{images/icon-modality.png} X-ray & \includegraphics[height=0.8em]{images/icon-area.png} \textbf{Lung} & \includegraphics[height=0.8em]{images/icon-modality.png} \textbf{CT} & \includegraphics[height=0.8em]{images/icon-area.png} \textbf{Lung} & 49 & 57 & \ding{51} \\
\includegraphics[height=0.8em]{images/icon-modality_and_area.png} \textbf{FP - Fundus} & \includegraphics[height=0.8em]{images/icon-task.png} Diseases & \includegraphics[height=0.8em]{images/icon-modality_and_area.png} Mic - Cell & \includegraphics[height=0.8em]{images/icon-task.png} \textbf{Level} & \includegraphics[height=0.8em]{images/icon-modality_and_area.png} \textbf{FP - Fundus} & \includegraphics[height=0.8em]{images/icon-task.png} \textbf{Level} & 55 & 61 & \ding{51} \\
\includegraphics[height=0.8em]{images/icon-modality_and_area.png} \textbf{Mic - Cell} & \includegraphics[height=0.8em]{images/icon-task.png} Recognition & \includegraphics[height=0.8em]{images/icon-modality_and_area.png} FP - Fundus & \includegraphics[height=0.8em]{images/icon-task.png} \textbf{Level} & \includegraphics[height=0.8em]{images/icon-modality_and_area.png} \textbf{Mic - Cell} & \includegraphics[height=0.8em]{images/icon-task.png} \textbf{Level} & 10 & 32 & \ding{51} \\
\bottomrule
\end{tabular}
}
\caption{Result of Llama-3.2-Vision on selected classification datasets in Med-MAT. \ding{51} in \textit{CG Helps} indicates successful generalization, while \ding{55} denotes failure.}
\label{tab:llama-vision-9b-exp}
\vspace{-1mm}
\end{table*}

\paragraph{Analysis of the results} The results in Table \ref{tab:moreatt-1} and \ref{tab:moreatt-2} indicate that the two new attributes show data leakage due to subtle visual differences in corresponding images (e.g., COVID-19 and pneumonia have similar features). Importantly, the MLLM trained with CG combinations still shows improvements on downstream tasks, confirming that our approach remains valid for new attributes.

\paragraph{Reason to choose the existing three attributes (MAT-Triplet: Modality, Area, Task)} We have considered additional categories such as age, gender, and finer disease classification, but we ultimately chose to focus on the MAT-Triplet categories for the following reasons.
\begin{itemize}
    \item \textbf{The boundaries between MAT-Triplet (Modality, Area, Task) are clear.} Different modalities and areas correspond to distinct imaging methods and body areas, leading to significant differences between images; different tasks also require the MLLM to extract specific information, demanding varied understanding of the images.
    \item \textbf{All datasets can be annotated using MAT-Triplet (Modality, Area, Task) easily.} Other medical labels, such as gender and age, are only available in a small portion of datasets and are not suitable for large-scale annotation.
    \item \textbf{Similar categorization strategies have been adopted in previous studies.}
\end{itemize}


\begin{table*}[ht!]
\setlength{\tabcolsep}{2pt}
\centering
\small
\resizebox{0.87\textwidth}{!}{
\begin{tabular}{ll|ll|ll|ccc}
\toprule
\multicolumn{4}{c}{\textbf{Related Combination}} &  \multicolumn{2}{c}{\textbf{Target Subset}} & \textbf{Baseline} & \textbf{Trained} & \textbf{CG Helps} \\
\midrule
\includegraphics[height=0.8em]{images/icon-area.png} \textbf{Lung} & \includegraphics[height=0.8em]{images/icon-task.png} Lung Det & \includegraphics[height=0.8em]{images/icon-area.png} Bones & \includegraphics[height=0.8em]{images/icon-task.png} \textbf{Diseases} & \includegraphics[height=0.8em]{images/icon-area.png} \textbf{Lung} & \includegraphics[height=0.8em]{images/icon-task.png} \textbf{Diseases} & 49 & 52  & \ding{51} \\
\includegraphics[height=0.8em]{images/icon-area.png} \textbf{Lung} & \includegraphics[height=0.8em]{images/icon-task.png} Lung Det & \includegraphics[height=0.8em]{images/icon-area.png} Breast & \includegraphics[height=0.8em]{images/icon-task.png} \textbf{Diseases} & \includegraphics[height=0.8em]{images/icon-area.png} \textbf{Lung} & \includegraphics[height=0.8em]{images/icon-task.png} \textbf{Diseases} & 49 & 54 & \ding{51} \\
\includegraphics[height=0.8em]{images/icon-area.png} \textbf{Bones} & 
 \includegraphics[height=0.8em]{images/icon-task.png} Spinal Error Det & \includegraphics[height=0.8em]{images/icon-area.png} Breast & \includegraphics[height=0.8em]{images/icon-task.png} \textbf{Diseases} & \includegraphics[height=0.8em]{images/icon-area.png} \textbf{Bones} & \includegraphics[height=0.8em]{images/icon-task.png} \textbf{Diseases} & 20 & 30 & \ding{51} \\
\includegraphics[height=0.8em]{images/icon-area.png} \textbf{Bones} & \includegraphics[height=0.8em]{images/icon-task.png} Spinal Error Det & \includegraphics[height=0.8em]{images/icon-area.png} Lung & \includegraphics[height=0.8em]{images/icon-task.png} \textbf{Diseases} & \includegraphics[height=0.8em]{images/icon-area.png} \textbf{Bones} & \includegraphics[height=0.8em]{images/icon-task.png} \textbf{Diseases} & 20 & 33 & \ding{51} \\
\midrule
\includegraphics[height=0.8em]{images/icon-modality.png} \textbf{End} & \includegraphics[height=0.8em]{images/icon-task.png} Level & \includegraphics[height=0.8em]{images/icon-modality.png} MRI & \includegraphics[height=0.8em]{images/icon-task.png} \textbf{Diseases Det} & \includegraphics[height=0.8em]{images/icon-modality.png} \textbf{End} & \includegraphics[height=0.8em]{images/icon-task.png} \textbf{Diseases} & 24 & 27 & \ding{51} \\
\includegraphics[height=0.8em]{images/icon-modality.png} \textbf{X-ray} & \includegraphics[height=0.8em]{images/icon-task.png} Lung Det & \includegraphics[height=0.8em]{images/icon-modality.png} CT & \includegraphics[height=0.8em]{images/icon-task.png} \textbf{COVID} & \includegraphics[height=0.8em]{images/icon-modality.png} \textbf{X-ray} & \includegraphics[height=0.8em]{images/icon-task.png} \textbf{COVID} & 23 & 26 & \ding{51} \\
\midrule
\includegraphics[height=0.8em]{images/icon-modality_and_area.png} \textbf{Der} - \textbf{Skin} & \includegraphics[height=0.8em]{images/icon-task.png} Cancer Det & \includegraphics[height=0.8em]{images/icon-modality_and_area.png} FP - Fundus & \includegraphics[height=0.8em]{images/icon-task.png} \textbf{Diseases} & \includegraphics[height=0.8em]{images/icon-modality_and_area.png} \textbf{Der} - \textbf{Skin} & \includegraphics[height=0.8em]{images/icon-task.png} \textbf{Diseases} & 24 & 29 & \ding{51} \\
\includegraphics[height=0.8em]{images/icon-modality_and_area.png} \textbf{Mic} - \textbf{Cell} & \includegraphics[height=0.8em]{images/icon-task.png} Cancer Det & \includegraphics[height=0.8em]{images/icon-modality_and_area.png} CT - Kidney & \includegraphics[height=0.8em]{images/icon-task.png} \textbf{Diseases} & \includegraphics[height=0.8em]{images/icon-modality_and_area.png} \textbf{Mic} - \textbf{Cell} & \includegraphics[height=0.8em]{images/icon-task.png} \textbf{Diseases} & 24 & 26 & \ding{51} \\
\bottomrule
\end{tabular}
}
\caption{Result of NEXT-Chat on CG by using detection and classification tasks to generalize classification Target dataset. Generalization results on classification datasets: \textit{Related Combination} is the training set, \textit{Target Subset} is the goal. Baseline and Trained represent the model's accuracy without training and trained on related data, respectively. \ding{51} in \textit{CG Helps} indicates successful generalization, while \ding{55} denotes failure.}
\label{tab:CG-det-next}
\end{table*}

\begin{table*}[ht!]
\setlength{\tabcolsep}{2pt}
\centering
\small
\resizebox{0.87\textwidth}{!}{
\begin{tabular}{ll|ll|ll|ccc}
\toprule
\multicolumn{4}{c}{\textbf{Related Combination}} &  \multicolumn{2}{c}{\textbf{Target Subset}} & \textbf{Baseline} & \textbf{Trained} & \textbf{CG Helps} \\
\midrule
\includegraphics[height=0.8em]{images/icon-area.png} \textbf{Lung} & \includegraphics[height=0.8em]{images/icon-task.png} Lung Det & \includegraphics[height=0.8em]{images/icon-area.png} Bones & \includegraphics[height=0.8em]{images/icon-task.png} \textbf{Diseases} & \includegraphics[height=0.8em]{images/icon-area.png} \textbf{Lung} & \includegraphics[height=0.8em]{images/icon-task.png} \textbf{Diseases} & 41 & 47  & \ding{51} \\
\includegraphics[height=0.8em]{images/icon-area.png} \textbf{Lung} & \includegraphics[height=0.8em]{images/icon-task.png} Lung Det & \includegraphics[height=0.8em]{images/icon-area.png} Breast & \includegraphics[height=0.8em]{images/icon-task.png} \textbf{Diseases} & \includegraphics[height=0.8em]{images/icon-area.png} \textbf{Lung} & \includegraphics[height=0.8em]{images/icon-task.png} \textbf{Diseases} & 41 & 49 & \ding{51} \\
\includegraphics[height=0.8em]{images/icon-area.png} \textbf{Bones} & \includegraphics[height=0.8em]{images/icon-task.png} Spinal Error Det & \includegraphics[height=0.8em]{images/icon-area.png} Breast & \includegraphics[height=0.8em]{images/icon-task.png} \textbf{Diseases} & \includegraphics[height=0.8em]{images/icon-area.png} \textbf{Bones} & \includegraphics[height=0.8em]{images/icon-task.png} \textbf{Diseases} & 31 & 35 & \ding{51} \\
\includegraphics[height=0.8em]{images/icon-area.png} \textbf{Bones} & \includegraphics[height=0.8em]{images/icon-task.png} Spinal Error Det & \includegraphics[height=0.8em]{images/icon-area.png} Lung & \includegraphics[height=0.8em]{images/icon-task.png} \textbf{Diseases} & \includegraphics[height=0.8em]{images/icon-area.png} \textbf{Bones} & \includegraphics[height=0.8em]{images/icon-task.png} \textbf{Diseases} & 31 & 37 & \ding{51} \\
\midrule
\includegraphics[height=0.8em]{images/icon-modality.png} \textbf{End} & \includegraphics[height=0.8em]{images/icon-task.png} Level & \includegraphics[height=0.8em]{images/icon-modality.png} MRI & \includegraphics[height=0.8em]{images/icon-task.png} \textbf{Diseases Det} & \includegraphics[height=0.8em]{images/icon-modality.png} \textbf{End} & \includegraphics[height=0.8em]{images/icon-task.png} \textbf{Diseases} & 24 & 26 & \ding{51} \\
\includegraphics[height=0.8em]{images/icon-modality.png} \textbf{X-ray} & \includegraphics[height=0.8em]{images/icon-task.png} Lung Det & \includegraphics[height=0.8em]{images/icon-modality.png} CT & \includegraphics[height=0.8em]{images/icon-task.png} \textbf{COVID} & \includegraphics[height=0.8em]{images/icon-modality.png} \textbf{X-ray} & \includegraphics[height=0.8em]{images/icon-task.png} \textbf{COVID} & 22 & 23 & \ding{51} \\
\midrule
\includegraphics[height=0.8em]{images/icon-modality_and_area.png} \textbf{Der} - \textbf{Skin} & \includegraphics[height=0.8em]{images/icon-task.png} Cancer Det & \includegraphics[height=0.8em]{images/icon-modality_and_area.png} FP - Fundus & \includegraphics[height=0.8em]{images/icon-task.png} \textbf{Diseases} & \includegraphics[height=0.8em]{images/icon-modality_and_area.png} \textbf{Der} - \textbf{Skin} & \includegraphics[height=0.8em]{images/icon-task.png} \textbf{Diseases} & 27 & 30 & \ding{51} \\
\includegraphics[height=0.8em]{images/icon-modality_and_area.png} \textbf{Mic} - \textbf{Cell} & \includegraphics[height=0.8em]{images/icon-task.png} Cancer Det & \includegraphics[height=0.8em]{images/icon-modality_and_area.png} CT - Kidney & \includegraphics[height=0.8em]{images/icon-task.png} \textbf{Diseases} & \includegraphics[height=0.8em]{images/icon-modality_and_area.png} \textbf{Mic} - \textbf{Cell} & \includegraphics[height=0.8em]{images/icon-task.png} \textbf{Diseases} & 20 & 24 & \ding{51} \\
\bottomrule
\end{tabular}
}
\caption{Result of MiniGPT-v2 on CG by using detection and classification tasks to generalize classification Target dataset. Generalization results on classification datasets: \textit{Related Combination} is the training set, \textit{Target Subset} is the goal. Baseline and Trained represent the model's accuracy without training and trained on related data, respectively. \ding{51} in \textit{CG Helps} indicates successful generalization, while \ding{55} denotes failure.}
\label{tab:CG-det-mini}
\end{table*}


\subsection{Details of Section \ref{sec: CG-on-other-backbones}: Exploring CG on different MLLM Backbones}
\label{app:RQ5}
To ensure the experiment results are not influenced by the model choice, we also tested several other models on some subsets of Med-MAT and observed similar results.



\textbf{Selection Strategy}: For testing, some generalized combinations were selected from classification tasks~\ref{tab:Compositional Generalization}. Using a random seed of 42, we shuffled each Direction Type’s combinations and selected the first two compositions as test data.

\textbf{Experimental Setup}: We conducted experiments to evaluate the compatibility of CG across different backbone architectures. We selected two MLLMs with representative architectures, namely Qwen2-VL-7B-Instruct~\cite{wang2024qwen2vlenhancingvisionlanguagemodels} and Llama-3.2-11B-Vision-Instruct~\cite{llama3.2}, to assess the performance of CG on these models. Each experiment involved full-parameter fine-tuning of all models over 5 epochs, utilizing 8 A800 (80GB) GPUs. The training was performed with a batch size of 32 and a learning rate set to 2e-6, ensuring that all parameters were updated to optimize the model performance.



\begin{table*}[ht!]
\setlength{\tabcolsep}{2pt}
\centering
\small
\resizebox{0.88\textwidth}{!}{
\begin{tabular}{ll|ll|ll|r}
\toprule
\multicolumn{4}{c}{\textbf{Related Combination}} &  \multicolumn{2}{c}{\textbf{Target Subset}} & \textbf{HuatuoGPT} \\
\midrule
\includegraphics[height=0.8em]{images/icon-area.png} \textbf{Bones} & \includegraphics[height=0.8em]{images/icon-task.png} State & \includegraphics[height=0.8em]{images/icon-area.png} Breast & \includegraphics[height=0.8em]{images/icon-task.png} \textbf{Diseases} & \includegraphics[height=0.8em]{images/icon-area.png} \textbf{Bones} & \includegraphics[height=0.8em]{images/icon-task.png} \textbf{Diseases} & $+6.12$ \\
\includegraphics[height=0.8em]{images/icon-area.png} \textbf{Lung} & \includegraphics[height=0.8em]{images/icon-task.png} COVID & \includegraphics[height=0.8em]{images/icon-area.png} Bones & \includegraphics[height=0.8em]{images/icon-task.png} \textbf{Diseases} & \includegraphics[height=0.8em]{images/icon-area.png} \textbf{Lung} & \includegraphics[height=0.8em]{images/icon-task.png} \textbf{Diseases} & $+15.00$ \\
\includegraphics[height=0.8em]{images/icon-modality.png} \textbf{X-ray} & \includegraphics[height=0.8em]{images/icon-task.png} Diseases & \includegraphics[height=0.8em]{images/icon-modality.png} CT & \includegraphics[height=0.8em]{images/icon-task.png} \textbf{COVID} & \includegraphics[height=0.8em]{images/icon-modality.png} \textbf{X-ray} & \includegraphics[height=0.8em]{images/icon-task.png} \textbf{COVID} & $+38.00$ \\
\includegraphics[height=0.8em]{images/icon-modality.png} \textbf{X-ray} & \includegraphics[height=0.8em]{images/icon-task.png} Diseases & \includegraphics[height=0.8em]{images/icon-modality.png} CT & \includegraphics[height=0.8em]{images/icon-task.png} \textbf{State} & \includegraphics[height=0.8em]{images/icon-modality.png} \textbf{X-ray} &  \includegraphics[height=0.8em]{images/icon-task.png} \textbf{State} & $+40.67$ \\
\includegraphics[height=0.8em]{images/icon-modality.png} \textbf{CT} & \includegraphics[height=0.8em]{images/icon-area.png} Brain & \includegraphics[height=0.8em]{images/icon-modality.png} X-ray & \includegraphics[height=0.8em]{images/icon-area.png} \textbf{Lung} & \includegraphics[height=0.8em]{images/icon-modality.png} \textbf{CT} & \includegraphics[height=0.8em]{images/icon-area.png} \textbf{Lung} & $+1.5$ \\
\includegraphics[height=0.8em]{images/icon-modality.png} \textbf{CT} & \includegraphics[height=0.8em]{images/icon-area.png} Brain & \includegraphics[height=0.8em]{images/icon-modality.png} X-ray & \includegraphics[height=0.8em]{images/icon-area.png} \textbf{Lung} & \includegraphics[height=0.8em]{images/icon-modality.png} \textbf{CT} & \includegraphics[height=0.8em]{images/icon-area.png} \textbf{Lung} & $+18.00$ \\
\includegraphics[height=0.8em]{images/icon-modality_and_area.png} \textbf{FP - Fundus} & \includegraphics[height=0.8em]{images/icon-task.png} Diseases & \includegraphics[height=0.8em]{images/icon-modality_and_area.png} Mic - Cell & \includegraphics[height=0.8em]{images/icon-task.png} \textbf{Level} & \includegraphics[height=0.8em]{images/icon-modality_and_area.png} \textbf{FP - Fundus} & \includegraphics[height=0.8em]{images/icon-task.png} \textbf{Level} & $+12.12$ \\
\includegraphics[height=0.8em]{images/icon-modality_and_area.png} \textbf{Mic - Cell} & \includegraphics[height=0.8em]{images/icon-task.png} Recognition & \includegraphics[height=0.8em]{images/icon-modality_and_area.png} FP - Fundus & \includegraphics[height=0.8em]{images/icon-task.png} \textbf{Level} & \includegraphics[height=0.8em]{images/icon-modality_and_area.png} \textbf{Mic - Cell} & \includegraphics[height=0.8em]{images/icon-task.png} \textbf{Level} & $+10.50$ \\
\bottomrule
\end{tabular}
}
\caption{Result of HuatuoGPT-Vision on selected classification datasets in Med-MAT. \textit{HuatuoGPT} represent the accuracy(\%) gains the model achieved through CG.}
\label{tab:huatuogpt}
\vspace{-3mm}
\end{table*}

\subsection{Details of Section \ref{sec: cg-across-location-and-classification}: Exploring CG across Detection and Classification}
\label{app:RQ4}


\textbf{Experimental Setup}: We conducted generalization experiments for detection and classification. Specifically, we performed generalization validation on Next-Chat~\cite{zhang2023next} and MiniGPT-v2~\cite{chen2023minigpt}. Next-Chat models the bounding box as an embedding and utilizes a decoder for decoding, while MiniGPT-v2 treats the bounding box as a text token, which are common approaches used by existing MLLM implementations for detection. By conducting CG validation using distinct bounding box modeling methods, we further demonstrate the broad applicability of the CG approach. Each experiment was conducted on 8 A800 (80GB) GPUs.

The two backbones were trained separately in this experiment. For Next-Chat, we directly trained the model in its second training stage and fine-tuned it for 2 epochs with a learning rate of 2e-5, keeping all other training parameters at their default settings. Similarly, for MiniGPT-v2, we trained the backbone model from the second stage, starting with a learning rate of 2e-5 and gradually reducing it to 2e-6 over 3 epochs.



\subsection{CG with Medical Multimodal LLM}

In previous experiments, general MLLMs are selected to prevent the MLLM's inherent medical knowledge from affecting CG results. Our experiments focus on how MLLMs leverage CG to interpret unseen medical images. If the model has learned some fundamental elements of the \textit{Target} data, it would compromise the fairness of the experiments.

To demonstrate that our results still work on medical LLMs, we employed the same data combinations from Section \ref{sec: CG-on-other-backbones} to investigate CG on medical MLLMs (we selected HuatuoGPT-Vision as the baseline).

The results in Table \ref{tab:huatuogpt} demonstrate that \textbf{the medical-expert MLLM can still leverage CG to enhance their performance on novel tasks}, further supporting the validity and consistency of our findings.


\section{The Dataset: Med-MAT}
\label{app:dataset}

This section provides an overview of Med-MAT. First, a detailed explanation of MAT-Triplet will be presented in \ref{app:mat-triplet}. Next, the methods for constructing the QA formatting will be discussed in \ref{app:construction}. Finally, the data composition details and open-source specification will be provided in \ref{app:composition}.



\subsection{Details of MAT-Triplet}
\label{app:mat-triplet}

\textbf{MAT-Triplet} stands for \textbf{M}edical Modality, \textbf{A}natomical Area, and Medical \textbf{T}ask. We define all samples in Med-MAT using these three components and integrate datasets with identical triplets into subsets.

\textbf{Medical Modality} refers to different types of techniques or methods used in medical imaging or data acquisition. Each modality is designed to present the human body’s structures or pathological features in unique ways, providing auxiliary support for clinical diagnosis and treatment. Most modalities exhibit significant visual differences, making them easily distinguishable. Med-MAT encompasses 11 modalities, including common ones such as Computed Tomography (CT), Magnetic Resonance Imaging (MRI), X-ray, Fundus Photography (FP), Endoscopy (End), Optical Coherence Tomography (OCT), and Ultrasound (US), as well as rare and specialized modalities like Colonoscopy (Co), Dermoscopy (Der), Digital Pathology (DP), and Microscopy (Mic).

\textbf{Anatomical Area} refers to specific anatomical structures or regions within the human body or other organisms, defined by distinct anatomical characteristics to describe various body parts, their functions, and relative positions. Med-MAT encompasses 14 anatomical areas, including the cervix, kidney, lung, brain, intestine, bladder, fundus, retina, breast, bones, and chest. To further facilitate data description, additional categories such as skin, mouth, and cell are included as specialized anatomical areas.

\textbf{Medical Task} refers to the specific detection task that needs to be performed on the dataset. Med-MAT includes 13 distinct tasks, with classification tasks encompassing Quality Identification (image quality analysis), COVID Diagnosis, Cancer Diagnosis (determining the presence of a specific disease), State (such as identifying brain hemorrhage), Level Identification (assessing disease severity), and Multiple Classification (classifying multiple diseases or cell types). Given the limited options of COVID Diagnosis and Cancer Diagnosis, these tasks can be interpreted as identifying whether a patient is in a diseased state. To enhance generalization and provide more diverse examples, these tasks are grouped under the broader category of State. In addition, we have 16 datasets defining segmentation or classification tasks with different objectives.


\subsection{QA construction method}
\label{app:construction}

A large amount of image-label datasets was collected to build the Med-MAT dataset. To ensure compatibility with MLLM training inputs and outputs, all data is transformed into a question-answering format. Questions are formulated based on modality, anatomical area, and medical task, with 6 question prompts applied to each subset.

The labels within each data subset will be clustered to prevent redundant definitions of the same condition. Then, all training set and test set will be converted into multiple-choice questions following the template in Table~\ref{box:QAtemplate}. Each question will have up to four options, with distractor options randomly selected from the corresponding subset.




\subsection{Data composition and Open-source Specification}
\label{app:composition}

Med-MAT is composed of multiple datasets. After being transformed into different QA formats, the new data is organized into several subsets to support generalization experiments in medical imaging. Table \ref{tab:trinity-set} shows all of our subset datasets, which are separated based on different combinations in MAT-Triplet. The specific MAT-Triplets are listed, along with the labels corresponding to the image-label datasets for each subset. Correspondingly, all the image-label datasets are also displayed in Table \ref{tab:medical-datasets1}, which includes their names, descriptions of the tasks performed, download links, and the level of accessibility.

All question-answering text datasets in Med-MAT will be publicly available. To accommodate varying access permissions, we will release datasets based on their respective licenses: openly accessible datasets will be directly available, while restricted datasets can be accessed by applying through the links provided in this paper. We hope this dataset will support and advance future generalization experiments on medical imaging.

\subsection{Data Sources and Distribution}

All Med-MAT data are sourced from public medical image challenges or widely used, high-impact datasets previously applied in deep learning training, ensuring reliable annotations. Before inclusion in Med-MAT, all datasets underwent label averaging where possible; test sets, in particular, were strictly balanced to ensure accuracy reliably reflects model performance. Each Med-MAT training subset contains 3,000 samples, while test sets maximize size under label balance constraints.

\section{Bad cases analysis and solutions}

\subsection{Bad case analysis}

Some \textit{Trained} models show minimal gains or even performance declines in Table \ref{tab:Compositional Generalization}, with classification accuracy lower than either the \textit{Baseline} or \textit{Baseline+}. After a thorough examination, we found that these Target datasets require more fine-grained medical condition classification. Beyond disease presence, they need detailed assessments, such as severity grading (e.g., bone age estimation, cancer staging) or distinguishing similar conditions (e.g., differentiating COVID-19 from pneumonia).

\begin{itemize}
    \item The \textit{Related} combinations lack suitable fundamental elements: For CG, the training data must include the \textit{Target} task's core elements. Here, we use other "level classification/grading" tasks for generalization, but their criteria differ significantly, misaligning with the \textit{Target} task's needs.
    \item Without defined grading standards, MLLMs lacking relevant knowledge can't perform fine-grained tasks: Tasks like bone age assessment and cancer staging vary by criteria, and without this knowledge, MLLMs can't accurately classify them.
\end{itemize}

\subsection{Possible solutions}


\paragraph{Few-shot prompting} As we illustrated before, most of the bad cases involve fine-grained tasks needing specialized knowledge. So, in order to minimize the effect of a lack of relevant knowledge, we also conducted few-shot experiments to add some target images in the prompts. Subset \textit{X-ray, Lung, Normal-COVID-Pneumonia} was chosen for its simple structure, with LLaVA as the baseline. We randomly sampled n images per label for n-shot inference and repeated each experiment 3 times.

\begin{table}[ht!]
\setlength{\tabcolsep}{2pt}
\centering
\small
\resizebox{0.48\textwidth}{!}{
\begin{tabular}{l|l|lll}
\toprule
\textbf{Model} & \textbf{0-shot} & \textbf{2-shot} & \textbf{3-shot} & \textbf{4-shot} \\
\midrule
LLaVA & $30.00$ & $\textbf{28.83} \pm 0.85$ & $29.33 \pm 1.25$ & $29.83 \pm 1.31$ \\
LLaVA + CG & $28.00$ & $28.67 \pm 0.94$ & $\textbf{37.00} \pm 0.82$ & $\textbf{36.67} \pm 0.47$ \\
\bottomrule
\end{tabular}
}
\caption{Results of Few-shot prompting.}
\label{tab:badcase--few-shot}
\vspace{-2mm}
\end{table}

The results in Table~\ref{tab:badcase--few-shot} demonstrate that training with CG combinations can improve the few-shot performance of MLLMs on downstream tasks, even when direct CG generalization is not effective.

\paragraph{Adding some \textit{Target} data in training} As described in Section \ref{sec:ia-lowdata}, we selected cases where CG alone couldn't achieve satisfactory results and augmented their training sets with target data. The results in this section indicate that while CG may not directly enhance generalization, it accelerates the model’s adaptation to downstream tasks.

\begin{figure*}[ht!]
\begin{AIbox}{Multiple-choice Questions Template}
<question>

A. <option\_1>

B. <option\_2>

C. <option\_3>

D. <option\_4>

Answer with the option’s letter from the given choices directly.
\end{AIbox}
\caption{The Template of multiple-choice questions.}
\label{box:QAtemplate}
\end{figure*}

\begin{figure*}[ht!]
    \centering
    \includegraphics[width=\textwidth]{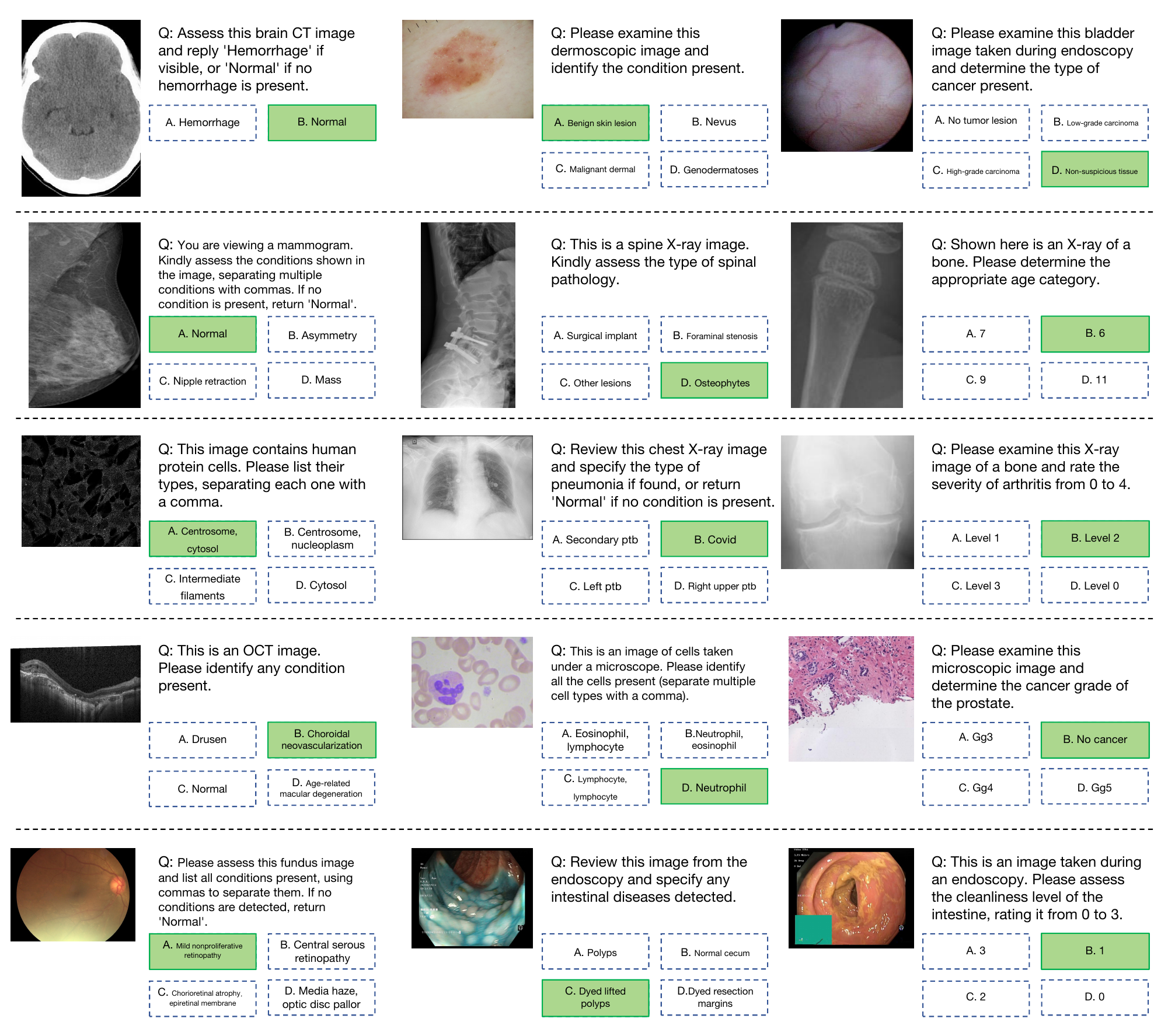}
    \caption{Illustration of diverse samples with varying numbers of candidate options in the Med-MAT dataset.}
    \label{fig:med-mat-show-examples}
\end{figure*}

\begin{table*}[ht!]
\setlength{\tabcolsep}{2pt}
\centering
\small
\begin{tabular}{l|ll|l|l}
\toprule
\textbf{Subset No.} & \textbf{Modality} & \textbf{Anatomical Area} & \textbf{Task} & \textbf{Datasets No.} \\
\midrule
\rowcolor{customblue!20} 01 & Co & Cervix & Cervical Picture Quality Evaluation & 1 \\
\rowcolor{customblue!20} 02 & CT & Kidney & Kidney Diseases Classification & 2 \\
\rowcolor{customblue!20} 03 & CT & Lung & COVID-19 Classification & 3,4,6 \\
\rowcolor{customblue!20} 04 & CT & Lung & Lung Cancer Classification & 5 \\
\rowcolor{customblue!20} 05 & CT & Brain & Brain Hemorrhage Classification & 7 \\
\rowcolor{customblue!20} 06 & CT & Brain & Brain Cancer Classification & 8 \\
\rowcolor{customblue!20} 07 & Der & Skin & Melanoma Type Classification & 10 \\
\rowcolor{customblue!20} 08 & Der & Skin & Skin Diseases Classification & 9, 11-15, 71, 72, 74 \\
\rowcolor{customblue!20} 09 & DP & Mouth & Teeth Condition Classification & 16 \\
\rowcolor{customblue!20} 10 & DP & Mouth & Oral Cancer Classification & 17 \\
\rowcolor{customblue!20} 11 & End & Intestine & Intestine Cleanliness Level & 18 \\
\rowcolor{customblue!20} 12 & End & Bladder & Cancer Degree Classification & 19 \\
\rowcolor{customblue!20} 13 & End & Intestine & Intestine Diseases Classification & 20 \\
\rowcolor{customblue!20} 14 & FP & Fundus & Eye Diseases Classification & 21-23, 26-28, 31, 32, 75 \\
\rowcolor{customblue!20} 15 & FP & Fundus & Multiple-labels Eye Diseases Classification & 24, 25, 68 \\
\rowcolor{customblue!20} 16 & FP & Fundus & Blindness Level & 29 \\
\rowcolor{customblue!20} 17 & FP & Fundus & Retinal Images Quality Evaluation & 30 \\
\rowcolor{customblue!20} 18 & Mic & Cell & Cell Type Classification & 33, 36-38, 39-41, 44, 65, 70 \\
\rowcolor{customblue!20} 19 & Mic & Cell & Prostate Cancer Degree Classification & 34\\
\rowcolor{customblue!20} 20 & Mic & Cell & Multiple-labels Blood Cell Classification & 35 \\
\rowcolor{customblue!20} 21 & Mic & Cell & Cancer Classification & 42, 67 \\
\rowcolor{customblue!20} 22 & MRI & Brain & Head Diseases Classification & 44, 45 \\
\rowcolor{customblue!20} 23 & OCT & Retina & Retina Diseases Classification & 46, 47 \\
\rowcolor{customblue!20} 24 & US & Breast & Breast Cancer Classification & 48 \\
\rowcolor{customblue!20} 25 & X-ray & Bones & Degree Classification of Knee & 49, 53 \\
\rowcolor{customblue!20} 26 & X-ray & Bones & Fractured Classification & 50, 51 \\
\rowcolor{customblue!20} 27 & X-ray & Bones & Vertebrae Diseases Classification & 52 \\
\rowcolor{customblue!20} 28 & X-ray & Lung & COVID-19 and Pneumonia Classification & 54-57, 60, 62, 81 \\
\rowcolor{customblue!20} 29 & X-ray & Breast & Breast Diseases Classification & 58, 78 \\
\rowcolor{customblue!20} 30 & X-ray & Lung & Tuberculosis Classification & 59, 79 \\
\rowcolor{customblue!20} 31 & X-ray & Chest & Multiple-labels Chest Classification & 61, 73, 76, 77, 80, 85, 87 \\
\rowcolor{customblue!20} 32 & X-ray & Brain & Tumor Classification & 63 \\
\rowcolor{customblue!20} 33 & Mic & Cell & Multi-labels Diseases & 84 \\
\rowcolor{customblue!20} 34 & FP & Fundus & Level Identification & 66 \\
\rowcolor{customblue!20} 35 & X-ray & Bones & Level Identification & 69 \\
\rowcolor{customblue!20} 36 & X-ray & Bones & Spinal lesion Classification & 86 \\
\rowcolor{customblue!20} 37 & X-ray & Breast & Multi-labels Diseases & 82 \\
\midrule
\rowcolor{customgreen!25} 38 & Der & Skin & Lesion Det/Seg & 88-91 \\
\rowcolor{customgreen!25} 39 & End & Intestine & PolyP Det/Seg & 92-93 \\
\rowcolor{customgreen!25} 40 & End & Intestine & Surgical Procedures Det/Seg & 94 \\
\rowcolor{customgreen!25} 41 & End & Intestine & Multi-labels Det/Seg & 95 \\
\rowcolor{customgreen!25} 42 & Mic & Cell & Cancer Cell Det/Seg & 96 \\
\rowcolor{customgreen!25} 43 & US & Chest & Cancer Det/Seg & 97 \\
\rowcolor{customgreen!25} 44 & US & Thyroid & Thyroid Nodule Region Det/Seg & 98 \\
\rowcolor{customgreen!25} 45 & MRI & Intestine & Multi-labels Det/Seg & 103 \\
\rowcolor{customgreen!25} 46 & MRI & Liver & Liver Det/Seg & 104, 105 \\
\rowcolor{customgreen!25} 47 & X-ray & Lung & Lung Det/Seg & 99 \\
\rowcolor{customgreen!25} 48 & X-ray & Lung & Pneumothorax Det/Seg & 106 \\
\rowcolor{customgreen!25} 49 & X-ray & Bones & Spinal Anomaly Det & 100 \\
\rowcolor{customgreen!25} 50 & X-ray & Chest & Multi-labels Det & 101, 102 \\
\rowcolor{customgreen!25} 51 & FP & Fundus & Vessel Seg & 107 \\
\rowcolor{customgreen!25} 52 & FP & Fundus & Optic Disc and Cup Seg & 108 \\
\rowcolor{customgreen!25} 53 & FP & Fundus & Optic Disc Seg & 109 \\
\bottomrule
\end{tabular}
\caption{The details of subset. In particular, \textbf{Co} stands for Colposcopy, \textbf{CT} represents Computed Tomography, \textbf{DP} refers to Digital Photography, \textbf{FP} is for Fundus Photography, \textbf{MRI} denotes Magnetic Resonance Imaging, \textbf{OCT} signifies Optical Coherence Tomography, \textbf{Der} refers to Dermoscopy, \textbf{End} stands for Endoscopy, \textbf{Mic} indicates Microscopy Images, and \textbf{US} represents Ultrasound. The \colorbox{customblue!20}{blue} section represents the classification dataset and the \colorbox{customgreen!35}{green} section represents the detection}
\label{tab:trinity-set}
\end{table*}

\begin{table*}[ht!]
\setlength{\tabcolsep}{2pt}
\centering
\small
\resizebox{\textwidth}{!}{
\begin{tabular}{l|l|l|l}
\toprule
\textbf{No.} & \textbf{Name} & \textbf{Description} & \textbf{Citation} \\
\midrule
1 & \href{https://www.kaggle.com/competitions/intel-mobileodt-cervical-cancer-screening/data}{Intel \& MobileODT Cervical Screening} & Cervix Type in Screening & \cite{intel-mobileodt-cervical-cancer-screening} \\
2 & \href{https://www.kaggle.com/datasets/nazmul0087/ct-kidney-dataset-normal-cyst-tumor-and-stone}{CT Kindney Dataset} & Normal or Cyst or Tumor & \cite{islam2022vision} \\
3 & \href{https://www.kaggle.com/datasets/plameneduardo/sarscov2-ctscan-dataset}{SARS-COV-2 Ct-Scan} & COVID19, Classification Dataset & \cite{2020SARS} \\
4 & \href{https://tianchi.aliyun.com/dataset/106604}{COVID CT COVID-CT} & COVID19, Classification Dataset & \cite{zhao2020COVID-CT-Dataset} \\
5 & \href{https://tianchi.aliyun.com/dataset/93929}{Chest CT-Scan} & Cancer Classification & \cite{tianchi93929} \\
6 & \href{https://tianchi.aliyun.com/dataset/93666}{COVID-19-CT SCAN IMAGES} & COVID19, Classification & \cite{COVID-19-CT} \\
7 & \href{https://www.kaggle.com/datasets/felipekitamura/head-ct-hemorrhage?select=labels.csv}{Head CT} & Head Hemorrhage & \cite{felipe_campos_kitamura_2018} \\
8 & \href{https://www.kaggle.com/datasets/trainingdatapro/computed-tomography-ct-of-the-brain}{CT of Brain} & Head Cancer & \cite{ctofbrain} \\
9 & \href{https://www.cs.rug.nl/~imaging/databases/melanoma_naevi/}{MED-NODE} & Melanoma or Naevus & \cite{giotis2015med} \\
10 & \href{https://challenge2020.isic-archive.com/}{ISIC 2020} & Melanoma, Benign or Malignant & \cite{rotemberg2021patient} \\
11 & \href{https://data.mendeley.com/datasets/zr7vgbcyr2/1}{PAD-UFES-20} & Skin Multi Classification & \cite{pacheco2020pad} \\
12 & \href{https://www.kaggle.com/datasets/arafathussain/monkeypox-skin-image-dataset-2022, https://www.heywhale.com/mw/dataset/62eb75d6fef0903951b1f199}{Web-scraped Skin Image} & Skin Desease Multi Classification & \cite{islam2022aweb} \\
13 & \href{https://www.kaggle.com/datasets/angelachristabel/isbi-2016?select=Training_GroundTruth.csv}{ISBI 2016} & Skin Lesion Classification & \cite{gutman2016skin} \\
14 & \href{https://www.kaggle.com/datasets/andrewmvd/isic-2019}{ISIC 2019} & Skin Desease Multi Classification & \cite{combalia2019bcn20000} \\
15 & \href{https://www.kaggle.com/datasets/nodoubttome/skin-cancer9-classesisic}{Skin Cancer ISIC} & Skin Cancer Multi Classification & \cite{skin-cancer-isic} \\
16 & \href{https://www.kaggle.com/datasets/salmansajid05/oral-diseases/data}{Dental Condition Dataset} & Teeth condition classification & \cite{oral-diseases} \\
17 & \href{https://www.kaggle.com/datasets/zaidpy/oral-cancer-dataset}{Oral Cancer Dataset} & Oral cancer Classification & \cite{oral-cancer-dataset} \\
18 & \href{https://datasets.simula.no/nerthus/}{The Nerthus Dataset} & Cleanliness level & \cite{Pogorelov:2017:NBP:3083187.3083216} \\
19 & \href{https://commons.datacite.org/doi.org/10.5281/zenodo.7741475}{Endoscopic Bladder Tissue} & Canser Degree Classification & \cite{lazo2023semi} \\
20 & \href{https://www.kaggle.com/datasets/meetnagadia/kvasir-dataset}{Kvasir} & Multi Disease Classification & \cite{Pogorelov:2017:KMI:3083187.3083212} \\
21 & \href{https://figshare.com/s/c2d31f850af14c5b5232}{ACRIMA} & Glaucoma & \cite{ovreiu2021deep} \\
22 & \href{https://www.kaggle.com/datasets/nurmukhammed7/augemnted-ocular-diseases}{Augemnted ocular diseases AOD} & Multi Classification of eye diseases & \cite{augemnted-ocular-diseases} \\
23 & \href{https://www.kaggle.com/datasets/linchundan/fundusimage1000}{JSIEC} & Multi Classification of eye diseases & \cite{cen2021automatic} \\
24 & \href{https://data.mendeley.com/datasets/pc4mb3h8hz/1}{Multi-Label Retinal Diseases} & Multi Classification of eye diseases & \cite{rodriguez2022multi} \\
25 & \href{https://github.com/openmedlab/Awesome-Medical-Dataset/blob/main/resources/RFMiD.md}{RFMiD 2.0} & Multi Classification of eye diseases & \cite{panchal2023retinal} \\
26 & \href{https://www.kaggle.com/datasets/nafin59/ocular-toxoplasmosis-fundus-images-dataset}{ToxoFundus(Data Processed Paper)} & Ocular toxoplasmosis & \cite{CARDOZO2023109056} \\
27 & \href{https://www.kaggle.com/datasets/nafin59/ocular-toxoplasmosis-fundus-images-dataset}{ToxoFundus(Data Raw 6class All)} & Ocular toxoplasmosis & \cite{CARDOZO2023109056} \\
28 & \href{https://www.kaggle.com/datasets/xiaoliang2121/adamdataset}{Adam dataset} & Age-related Macular Degeneration & \cite{adam-dataset} \\
29 & \href{https://www.kaggle.com/competitions/aptos2019-blindness-detection}{APTOS 2019 Blindness} & Blindness Level Identification & \cite{aptos-2019-blindness-detection} \\
30 & \href{https://www.kaggle.com/datasets/subhajournal/drimdb-diabetic-retinopathy-images-database}{DRIMDB} & Quality Testing of Retinal Images & \cite{inproceedings} \\
31 & \href{https://www.kaggle.com/datasets/sshikamaru/glaucoma-detection}{Glaucoma Detection} & Glaucoma Classification & \cite{glaucoma-detection} \\
32 & \href{https://zenodo.org/records/5793241}{AIROGS} & Glaucoma Classification & \cite{devente23airogs} \\
33 & \href{https://github.com/KaikaiZhao/HEp-2_cell_classification}{ICPR-HEp-2} & Multi Classification & \cite{qi2016hep} \\
34 & \href{https://data.mendeley.com/datasets/9xxm58dvs3/1}{SICAPv2} & Cancer Degree Classification & \cite{silva2020going} \\
35 & \href{https://www.kaggle.com/datasets/paultimothymooney/blood-cells}{Blood Cell Images} & Blood Cell Classificaion & \cite{blood-cell-images} \\
36 & \href{https://www.kaggle.com/datasets/ambarish/breakhis}{BreakHis} & Cell type and beginormag & \cite{break-his} \\
37 & \href{https://bupt-ai-cz.github.io/HSA-NRL/}{Chaoyang} & Multi Classification of pathologists & \cite{zhuhard} \\
38 & \href{https://data.mendeley.com/datasets/tt3yj2pf38/3}{HuSHeM} & Sperm Head Morphology Classificaion & \cite{shaker2018human} \\
39 & \href{https://www.kaggle.com/datasets/andrewmvd/bone-marrow-cell-classification}{Bone Marrow Cell Classification} & Bone Marrow Cell Classification & \cite{matek2021expert} \\
40 & \href{https://zenodo.org/records/1214456}{NCT-CRC-HE-100K} & Multi Classification & \cite{kather_jakob_nikolas_2018_1214456} \\
41 & \href{https://www.kaggle.com/datasets/andrewmvd/malignant-lymphoma-classification}{Malignant Lymphoma Classification} & Multi Classification & \cite{article1214456} \\
42 & \href{https://www.kaggle.com/c/histopathologic-cancer-detection/data}{Histopathologic Cancer Detection} & Cancer Classification & \cite{histopathologic-cancer-detection} \\
43 & \href{https://www.kaggle.com/datasets/xilezhu/lc25000}{LC25000} & Multi Classification of Lung and Colon & \cite{lc25000} \\
44 & \href{https://www.kaggle.com/datasets/fernando2rad/brain-tumor-mri-images-17-classes}{Brain Tumor 17 Classes} & Multi Classification & \cite{brain-tumor-mri-images-17-classes} \\
45 & \href{https://www.kaggle.com/datasets/masoudnickparvar/brain-tumor-mri-dataset}{Tumor Classification} & Pituitary or Glioma or Meningioma or Notumor & \cite{brain-tumor-mri-dataset} \\
46 & \href{https://www.kaggle.com/datasets/andrewmvd/malignant-lymphoma-classification}{Malignant Lymphoma Classification} & Multi Classification of eye diseases & \cite{articlemalignant} \\
47 & \href{https://www.kaggle.com/datasets/obulisainaren/retinal-oct-c8}{Retinal OCT-C8} & Multi Classification of eye diseases & \cite{9740985} \\
48 & \href{https://www.kaggle.com/datasets/sabahesaraki/breast-ultrasound-images-dataset}{BUSI} & Breast Cancer & \cite{al2020dataset} \\
49 & \href{https://data.mendeley.com/datasets/t9ndx37v5h/1}{Digital Knee X-Ray Images} & Degree Classification of Knee & \cite{gornale_shivanand_2020_41241861} \\
50 & \href{https://www.kaggle.com/datasets/preetviradiya/brian-tumor-dataset}{Bone Fracture Multi-Region X-ray Data} & Fractured Classification & \cite{msoud_nickparvar_2021} \\
51 & \href{https://www.kaggle.com/datasets/devbatrax/fracture-detection-using-x-ray-images}{Fracture detection} & Fractured Classification & \cite{fracture-detection-using-x-ray-images} \\
52 & \href{https://www.kaggle.com/datasets/yasserhessein/the-vertebrae-xray-images}{The vertebrae X-ray image} & Vertebrae & \cite{fraiwan2022using} \\
53 & \href{https://www.kaggle.com/datasets/shashwatwork/knee-osteoarthritis-dataset-with-severity}{Knee Osteoarthritis Dataset} & Knee Osteoarthritis with severity grading & \cite{chen2018knee} \\
54 & \href{https://lhncbc.nlm.nih.gov/LHC-downloads/downloads.html\#tuberculosis-image-data-sets}{Shenzhen Chest X-Ray Set} & COVID19, Classification Dataset & \cite{jaeger2014two} \\
55 & \href{https://data.mendeley.com/datasets/jctsfj2sfn/1}{Chest X-ray PD} & COVID and Pneumonia & \cite{asraf2021covid19} \\
56 & \href{https://www.heywhale.com/mw/dataset/6027caee891f960015c863d7/content}{COVID-19 CHEST X-RAY DATABASE} & COVID and Pneumonia & \cite{9144185} \\
57 & \href{https://github.com/ari-dasci/covidgr}{COVIDGR} & COVID19, Classification & \cite{tabik2020covidgr} \\
58 & \href{https://www.kaggle.com/datasets/kmader/mias-mammography}{MIAS} & Multi Classification of Breast & \cite{mias-mammography} \\
59 & \href{https://www.kaggle.com/datasets/tawsifurrahman/tuberculosis-tb-chest-xray-dataset}{Tuberculosis Chest X-Ray Database} & Tuberculosis & \cite{rahman2020reliable} \\
60 & \href{https://www.kaggle.com/datasets/andrewmvd/pediatric-pneumonia-chest-xray}{Pediatric Pneumonia Chest X-Ray} & Pneumonia Classification & \cite{kermany2018labeled} \\
\bottomrule
\end{tabular}
}
\caption{The details of the medical datasets are provided}
\label{tab:medical-datasets1}
\end{table*}

\begin{table*}[ht!]
\setlength{\tabcolsep}{2pt}
\centering
\small
\resizebox{\textwidth}{!}{
\begin{tabular}{l|l|l|l}
\toprule
\textbf{No.} & \textbf{Name} & \textbf{Description} & \textbf{Citation} \\
\midrule
61 & \href{https://www.kaggle.com/datasets/nih-chest-xrays/sample}{Random Sample of NIH Chest X-Ray Dataset} & Multi Classificaiton of Chest & \cite{wang2017chestx} \\
62 & \href{https://www.kaggle.com/datasets/praveengovi/coronahack-chest-xraydataset}{CoronaHack-Chest X-Ray} & Pnemonia Classifcition with Virus type & \cite{coronahack-chest-x-ray-dataset} \\
63 & \href{https://www.kaggle.com/datasets/preetviradiya/brian-tumor-dataset}{Brain Tumor Dataset} & Tumor Classification & \cite{brain-tumor-dataset} \\
64 & \href{https://github.com/mattgroh/fitzpatrick17k}{Fitzpatrick 17k (Nine Labels)} & Multi Classification & \cite{groh2021evaluating} \\
65 & \href{https://figshare.com/s/d6fb591f1beb4f8efa6f}{BioMediTech} & Multi Classification & \cite{nanni2016texture} \\
66 & \href{https://zenodo.org/records/4891308}{Diabetic retinopathy} & Diabetic Retinopathy Level & \cite{benitez2021dataset} \\
67 & \href{https://tianchi.aliyun.com/dataset/90101/notebook}{Leukemia} & Cancer Classification & \cite{codella2019skin} \\
68 & \href{https://odir2019.grand-challenge.org/introduction/}{ODIR-5K} & Multiple Labels Classification & \cite{odir2019} \\
69 & \href{https://aistudio.baidu.com/datasetdetail/69582/0}{Arthrosis} & Bone Age Classification & \cite{rus-chn} \\
70 & \href{https://bupt-ai-cz.github.io/HSA-NRL/}{HSA-NRL} & Multi Classification of pathologists & \cite{zhu2021hard} \\
71 & \href{https://challenge.isic-archive.com/data/\#2018}{ISIC 2018 (Task 3)} & Multi Classification & \cite{codella2019skin} \\
72 & \href{https://challenge.isic-archive.com/data/\#2018}{ISIC 2017 (Task 3)} & Multi Classification & \cite{codella2018skin} \\
73 & \href{https://opendatalab.com/OpenDataLab/ChestX-Det}{ChestX-Det} & Multi Classification & \cite{lian2021structure} \\
74 & \href{https://www.kaggle.com/datasets/nafin59/monkeypox-skin-lesion-dataset}{Monkeypox Skin Lesion Dataset} & Only Monkeypox & \cite{Nafisa2022} \\
75 & \href{https://www.kaggle.com/datasets/jr2ngb/cataractdataset}{Cataract Dataset} & Multi Classification & \cite{cataract-dataset-jr2ngp} \\
76 & \href{https://www.kaggle.com/datasets/raddar/chest-xrays-indiana-university?select=indiana_reports.csv}{ChestX-rays IndianaUniversity} & Multi-label Classification & \cite{chest-x-ray-indiana-university} \\
77 & \href{https://www.kaggle.com/datasets/willarevalo/chexpert-v10-small}{CheXpert v1.0 small} & Multi-label Classification & \cite{chexpert-v1-small} \\
78 & \href{https://www.kaggle.com/datasets/awsaf49/cbis-ddsm-breast-cancer-image-dataset}{CBIS-DDSM} & Multi Classification & \cite{lee2017curated} \\
79 & \href{https://www.kaggle.com/datasets/nurkaraca/nlm-montgomerycxrset}{NLM-TB} & Tuberculosis & \cite{nlm-montgomery-cxr-set} \\
80 & \href{https://nihcc.app.box.com/v/ChestXray-NIHCC/folder/36938765345}{ChestXray-NIHCC} & Multi-label Classification & \cite{chestxray-nihcc} \\
81 & \href{https://www.kaggle.com/datasets/andyczhao/covidx-cxr2}{COVIDx CXR-4} & COVID19, Classification & \cite{Wang2020} \\
82 & \href{https://www.kaggle.com/datasets/ssmann/vindr-mammo-dataset}{VinDr-Mammo} & Multi-label Classification & \cite{nguyen2023vindr} \\
83 & \href{https://data.mendeley.com/datasets/snkd93bnjr/1}{PBC dataset normal DIB} & Multi Classification & \cite{acevedo2020dataset} \\
84 & \href{https://www.kaggle.com/competitions/hpa-single-cell-image-classification/data?select=train.csv}{Human Protein Atlas} & Multi-label Classification & \cite{le2022analysis} \\
85 & \href{https://www.rsna.org/rsnai/ai-image-challenge/rsna-pneumonia-detection-challenge-2018}{RSNA Pneumonia Detection Challenge 2018} & Multi-label Classification & \cite{rsna-pneumonia-detection-challenge} \\
86 & \href{https://www.physionet.org/content/vindr-spinexr/1.0.0/}{VinDr-SpineXR} & Multi Classification of Bones Diseases & \cite{pham2021vindr} \\
87 & \href{https://physionet.org/content/vindr-pcxr/1.0.0/}{VinDr-PCXR} & Multi-label Classification & \cite{pham2022vindr} \\
88 & \href{https://paperswithcode.com/dataset/ph2}{PH2} & Melanoma Segmentation & \cite{mendoncca2015ph2} \\
89 & \href{https://www.kaggle.com/datasets/angelachristabel/isbi-2016?select=Training_GroundTruth.csv}{ISBI 2016 (Task3B)} & Melanoma Segmentation & \cite{gutman2016skin} \\
90 & \href{https://challenge.isic-archive.com/data/\#2018}{ISIC 2016 (Task 1)} & Melanoma Segmentation & \cite{gutman2016skin} \\
91 & \href{https://challenge.isic-archive.com/data/\#2018}{ISIC 2017} & Melanoma Segmentation & \cite{codella2018skin} \\
92 & \href{https://polyp.grand-challenge.org/CVCClinicDB/}{CVC-ClinicDB} & Polyp Segmentation & \cite{bernal2015wm} \\
93 & \href{https://datasets.simula.no/kvasir-seg/, https://github.com/DebeshJha/2020-MediaEval-Medico-polyp-segmentation/tree/master}{Kvasir-SEG} & Polyp segmentation & \cite{jha2020kvasir} \\
94 & \href{https://www.kaggle.com/datasets/salmanmaq/m2caiseg}{m2caiseg} & Surgical Instrument Segmentation & \cite{maqbool2020m2caiseg} \\
95 & \href{https://edd2020.grand-challenge.org/Data/}{EDD 2020} & Multiple Diseases Segmentation in Intestine & \cite{EDD2020Challenge} \\
96 & \href{https://data.mendeley.com/datasets/9xxm58dvs3/1}{SICAPv2} & Cancer Cells Segmentation & \cite{silva2020going} \\
97 & \href{https://www.kaggle.com/datasets/sabahesaraki/breast-ultrasound-images-dataset}{BUSI} & Cancer Segmentation & \cite{breast-ultrasound-images-dataset} \\
98 & \href{https://github.com/haifangong/TRFE-Net-for-thyroid-nodule-segmentation}{TN3K} & Thyroid Nodule Segmentation & \cite{gong2022thyroid} \\
99 & \href{https://openi.nlm.nih.gov/imgs/collections/NLM-MontgomeryCXRSet.zip}{NLM-TB} & Lung Segmentation (With left or right) & \cite{gong2021multi} \\
100 & \href{https://www.physionet.org/content/vindr-spinexr/1.0.0/}{VinDr-SpineXR} & Spinal X-ray Anaomaly Detection & \cite{pham2021vindr} \\
101 & \href{https://physionet.org/content/vindr-pcxr/1.0.0/}{VinDr-PCXR} & Multiple Diseases Segmentation in Chest & \cite{pham2022vindr} \\
102 & \href{https://opendatalab.com/OpenDataLab/ChestX-Det}{ChestX-Det} & Multiple Diseases Segmentation in Chest & \cite{lian2021structure} \\
103 & \href{https://www.kaggle.com/competitions/uw-madison-gi-tract-image-segmentation/overview}{UW-Madison Gl Tract Image Segmentation} & Surgical Instrument Segmentation & \cite{lee2024dataset} \\
104 & \href{https://zenodo.org/records/7774566}{Duke Liver Dataset MRI v1} & Liver Segmentation & \cite{duke-liver} \\
105 & \href{https://zenodo.org/records/7774566}{Duke Liver Dataset MRI v2} & Liver Segmentation & \cite{duke-liver} \\
106 & \href{https://www.kaggle.com/c/siim-acr-pneumothorax-segmentation}{SIIM-ACR Pneumothorax Segmentation} & Pneumothorax Segmentation & \cite{siim-acr-pneumothorax-segmentation} \\
107 & \href{https://figshare.com/articles/figure/FIVES_A_Fundus_Image_Dataset_for_AI-based_Vessel_Segmentation/19688169/1?file=34969398}{FIVES} & Fundus Vascular Segmentation & \cite{jin2022fives} \\
108 & \href{https://github.com/miag-ull/rim-one-dl?tab=readme-ov-file}{RIM-ONE DL} & Optic Disc and Cup Segmentation & \cite{RIMONEDLImageAnalStereol2346} \\
109 & \href{https://ieee-dataport.org/documents/palm-pathologic-myopia-challenge}{PALM19} & Optic Disc Segmentation & \cite{55pk-8z03-19} \\
\bottomrule
\end{tabular}
}
\caption{Continued from Table~\ref{tab:medical-datasets1}.}
\label{tab:medical-datasets2}
\end{table*}


\end{CJK}
\end{document}